\documentclass{article}

\usepackage{microtype}
\usepackage{graphicx}
\usepackage{subcaption}
\usepackage{booktabs} % for professional tables

\usepackage{hyperref}

\usepackage[accepted]{icml2026}

\usepackage{amsmath}
\usepackage{amssymb}
\usepackage{mathtools}
\usepackage{amsthm}
\usepackage{multirow}
\usepackage{booktabs} % for professional tables
\usepackage{pifont}
\usepackage[table]{xcolor}

\usepackage[capitalize,noabbrev]{cleveref}

\theoremstyle{plain}
\newtheorem{theorem}{Theorem}[section]

\theoremstyle{definition}
\newtheorem{definition}[theorem]{Definition}

\theoremstyle{remark}
\newtheorem{remark}{Remark}

\usepackage[textsize=tiny]{todonotes}

\icmltitlerunning{Suppress and Diversify: Refining Robust Pathways for Corruption Robustness}

\begin{document}

\twocolumn[
  \icmltitle{Suppress and Diversify: Refining Robust Pathways for Corruption Robustness}



  \icmlsetsymbol{equal}{*}

  \begin{icmlauthorlist}
    \icmlauthor{Jiangang Yang}{equal,yyy}
    \icmlauthor{Wenhui Shi}{equal,yyy,comp}
    \icmlauthor{Xiaoran Xu}{comp}
    \icmlauthor{Wenyue Chong}{comp}
    \icmlauthor{Luqing Luo}{yyy}
    \icmlauthor{Jing Xing}{yyy}
    \icmlauthor{Jian Liu}{yyy}
    %\icmlauthor{Firstname8 Lastname8}{sch}
    %\icmlauthor{Firstname8 Lastname8}{yyy,comp}
  \end{icmlauthorlist}

  \icmlaffiliation{yyy}{Institute of Microelectronics, Chinese Academy of Sciences, Beijing, China}
  \icmlaffiliation{comp}{University of Chinese Academy of Science, Beijing, China}
  %\icmlaffiliation{sch}{School of ZZZ, Institute of WWW, Location, Country}

  \icmlcorrespondingauthor{Jiangang Yang}{yangjiangang@ime.ac.cn}
  \icmlcorrespondingauthor{Jian Liu}{liujian@ime.ac.cn}

  \icmlkeywords{Machine Learning, ICML}

  \vskip 0.3in
]



\printAffiliationsAndNotice{}  % no special notice (required even if empty)

\begin{abstract}
Model robustness against natural image corruptions is essential for safety-critical applications. While existing methods primarily focus on implicit representation learning, we provide the first systematic exploration of computational pathways to explicitly characterize internal robustness. We identify a progressive decay of robust features across network layers and establish a functional dependency between the prevalence of these features and model performance. To exploit these insights, we propose Suppress and Diversify (S\&D), a non-intrusive refinement approach that enhances robustness by dynamically selecting robust pathways and diversifying them through symmetry-preserving transformations. S\&D is architecture-agnostic, parameter-free, and incurs zero test-time overhead. Extensive evaluations across eight benchmarks demonstrate that S\&D consistently improves performance across multiple vision tasks, diverse backbones, and complex real-world scenarios, highlighting its broad efficacy and scalability. Code is available at \url{https://github.com/JGyoung-UCAS/suppress_and_diversify}.
\end{abstract}
%%%%%%%%%%%%%%%%%%%%%%%%%%%%%%%%%%%%%%%%%%%%%%%%%%%%%%%%%%%%%%%%%%%%%%%%%%%%%%%

\section{Introduction}
\label{sec:intro}
Recent years have witnessed the success of neural networks in various computer vision tasks. However, recent works have revealed that neural networks are vulnerable to image corruptions~\cite{hendrycks2019benchmarking}: image changes~(e.g., noise, blur, or low lighting) that degrade visual quality can lead neural networks to make errors such as misclassifying a bicycle as a motorbike. Such vulnerability inevitably transfers to downstream 2D and 3D perception tasks~\citep{michaelis2019benchmarking,kamann2021benchmarking,dong2023benchmarking,zeng2024benchmarking}, raising serious security concerns in safety-critical applications.
%%%%%%%%%%%%%%%%%%%%%%%%%%%%%%%%%
\begin{figure}[t]
    \centering
    \includegraphics[width=0.9\linewidth]{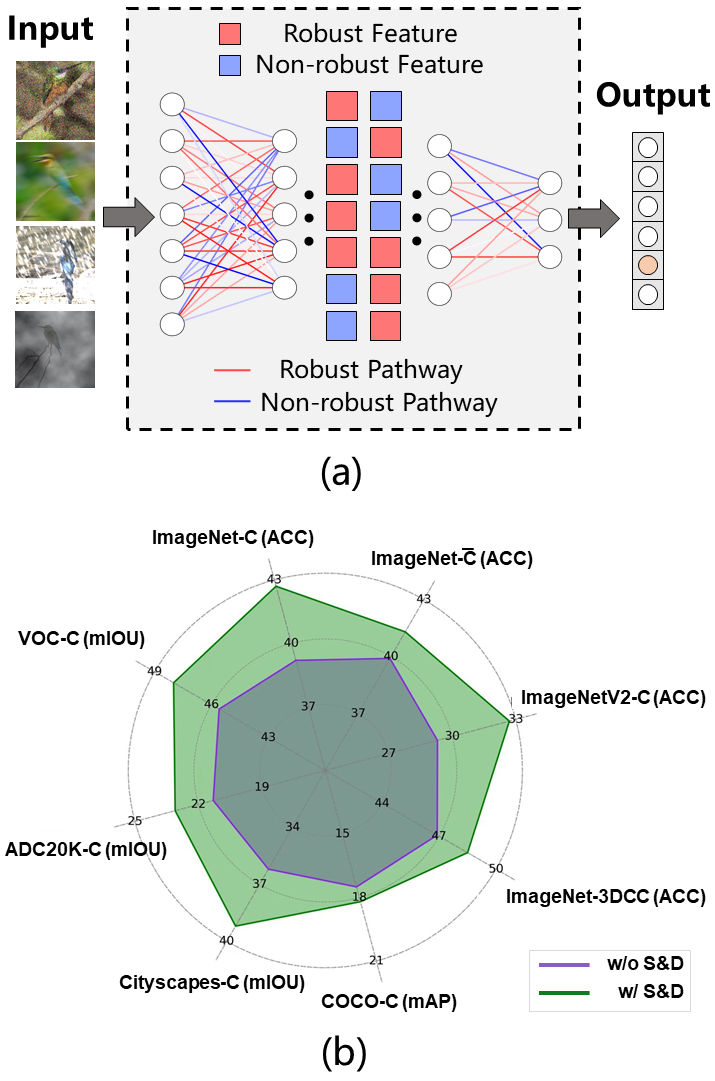}
    \caption{(a) The phenomenon of robust~(non-robust) features for image corruptions . (b) The performance gains of the proposed approach across various computer vision tasks.}
    \label{Figure 1}
\end{figure}

To solve this problem, prior approaches primarily focus on implicitly learning robust representations through data augmentation~\citep{hendrycks2019augmix,modas2022prime,qin2022understanding}, model regularization~\citep{zhang2018mixup,guo2023improving}, ensemble learning~\citep{saikia2021improving,diffenderfer2021winning}, and biological principles~\citep{teti2022lcanets,dapello2020simulating}. However, explicit modeling of robust features under image corruptions remains largely under-explored. While adversarial robustness research has extensively studied feature disentanglement~\citep{ilyas2019adversarial,tao2022can,kim2023feature,wang2024exploring}, these findings may not directly apply to natural corruptions, which exhibit global structural patterns and semantic shifts rather than local pixel-level noise. The nature of features under such structured distortions remains poorly understood, raising several critical questions: (a) \textit{Do neural networks similarly exhibit distinct non-robust features under image corruptions?} (b) \textit{Can we explicitly manipulate robust~(non-robust) features to enhance corruption robustness?}

%To solve this problem, many approaches focus on implicitly learning robust features from different perspectives of data augmentation~\citep{hendrycks2019augmix,modas2022prime,qin2022understanding}, model regularization~\citep{zhang2018mixup,guo2023improving}, ensemble learning~\citep{saikia2021improving,diffenderfer2021winning}, and biological principles~\citep {teti2022lcanets,dapello2020simulating}. However, explicit modeling of robust features for image corruptions has been less explored. In contrast, for adversarial attacks, another type of image perturbation that is human-made and imperceptible, the direct modeling of both robust and non-robust features has received widespread attention~\citep{ilyas2019adversarial,tao2022can}. Researchers have proposed disentangling robust and non-robust features to better understand and enhance adversarial robustness~\citep{kim2023feature,wang2024exploring}. Thus, several natural questions arise: (a) \textit{Do neural networks similarly exhibit non-robust features under image corruptions?} (b) \textit{Can we explicitly manipulate robust~(non-robust) features to enhance corruption robustness?}

To address these questions, we provide a systematic analysis to verify the existence and impact of non-robust features. We first formalize the robustness of computational pathways via the local Lipschitz constant~\citep{cisse2017parseval,fazlyab2019efficient}, revealing that robust and non-robust features are prevalent across the network. Our investigation identifies a progressive decay of robust features along the network depth, suggesting a non-homogeneous evolution of internal robustness. Furthermore, through linear probing and loss landscape analysis, we establish a functional dependency between robust feature prevalence and overall performance. These findings demonstrate that the aggregation of robust features dictates the robustness of downstream sub-networks and the full model under corruptions.

%To answer the questions above, we perform a comprehensive analysis to verify the existence of non-robust features. We begin by introducing a formal definition of individual feature robustness for image corruptions. Inspired by prior works~\citep{szegedy2014intriguing,weng2018evaluating}, we design a straightforward experiment via measuring the local Lipschitz constant~\citep{cisse2017parseval,fazlyab2019efficient} to demonstrate that robust~(non-robust) features are prevalent in the feature space of neural networks. In this way, we further investigate the collective behaviors of individual non-robust features. Our findings reveal a strong correlation between the aggregation of hidden features and robustness at the layer and sub-network levels, ultimately contributing to performance degradation under image corruptions.

Building on these insights, we propose Suppress and Diversify (S\&D), a non-intrusive refinement approach. The core principles are: (1) selecting pathways that generate robust features, and (2) diversifying them. Specifically, we introduce a Memory-aware Dynamic Selection Mechanism (MDSM) that maintains pathway groups in a memory bank, dynamically updated by a fitness score balancing feature invariance and discriminability. To prevent overfitting, a Structure-consistent Path Tweaking Strategy (SPTS) subsequently filters these pathways and generates augmented counterparts via symmetry-preserving transformations. This strategy enhances the diversity of robust representations while adhering to structural invariants. Finally, S\&D ensures the network is guided by a diverse set of robust computational pathways, significantly improving performance under corruptions with zero test-time overhead.

%Building on these insights, we propose Suppress\&Diversify (S\&D), a non-intrusive refinement approach. The core principles are: (1) selecting pathways that generate robust features, and (2) diversifying them. Specifically, a Memory-aware Dynamic Selection Mechanism~(MDSM) is introduced to guide the selection process. MDSM constructs a memory bank to store the group of pathways. To obtain the robust group, the memory bank is dynamically updated by a balanced metric that takes into account both the global robustness and discriminative ability of each group. Moreover, a Structure-consistent Path Tweaking Strategy~(SPTS) is designed to diversify the selected group. SPTS filters out non-robust pathways using a generic similarity metric. To enhance the diversity of robust features, SPTS stochastically tweaks the robust pathways to fill the gap within the group. Finally, S\&D refines learned robust features against image corruptions, thereby improving model robustness.

S\&D is a lightweight refinement requiring no inference-time overhead or structural modifications. We validate its efficacy across diverse benchmarks for classification, detection, and segmentation. S\&D consistently enhances the intrinsic robustness of various architectures and provides orthogonal gains when integrated with state-of-the-art methods. Our contributions are summarized as follows:
%S&D is a lightweight refinement with zero test-time overhead and requires no structural modifications to the backbone. We validate its efficacy through extensive evaluations across diverse corruption benchmarks, covering image classification, object detection, and semantic segmentation. The results demonstrate that S&D not only consistently enhances the intrinsic robustness of various architectures but also provides orthogonal performance gains when integrated with state-of-the-art robustness methods. Our contributions are summarized as follows:
%S\&D is easy to implement, with no run-time overheads and model structure modification. We conduct experiments across various corruption settings, including eight robustness benchmarks for image classification, object detection, and semantic segmentation. Extensive results demonstrate that S\&D not only improves robustness across different network backbones, but also boosts the robust performance combined with the current state-of-the-art methods. Our contributions are summarized as threefold:
\begin{itemize}
    \item We provide the first systematic exploration of non-robust features under image corruptions, revealing how the prevalence of these features determines the robustness across sub-networks and the overall model.
    \item We introduce S\&D, a novel training-time refinement that suppresses non-robust pathways and diversifies robust ones. S\&D is architecture-agnostic and parameter-free, enabling plug-and-play integration.
    \item Extensive evaluations across diverse benchmarks and vision tasks demonstrate that S\&D consistently enhances corruption robustness and OOD generalization.
\end{itemize}
\section{Related Work}
\label{sec:related_work}

%-------------------------------------------------------------------------
\subsection{Understanding Corruption Robustness}
There has been growing interest in understanding corruption robustness. Early work compares human and neural-network robustness under image corruptions~\cite{geirhos2018generalisation}. A major line of research explains robustness through model bias, especially texture/shape bias: texture bias plays a key role in corruption performance~\cite{geirhos2018imagenet}. Complementary perspectives analyze robustness in the Fourier domain by separating low- and high-frequency effects~\cite{yin2019fourier}, and connect robustness with decision-boundary thickness~\cite{yang2020boundary}. Large-scale evaluations show that several bias-based assumptions do not hold for network robustness~\cite{gavrikov2024can}. Closest to our focus, recent work implicitly models patch-induced non-robust features in Vision Transformers and shows that models exploit these patterns~\citep{qin2022understanding,guo2023improving}. However, no prior work explicitly models robust or non-robust features as a direct lens for understanding corruption robustness.

%There has been a surge of interest in understanding the corruption robustness of neural networks. An earlier study compares the robustness of humans and neural networks under image corruptions~\cite{geirhos2018generalisation}. One major direction is to rely on the assumption of texture/shape bias~(i.e., neural networks make predictions depending on image texture instead of shape information) to explain corruption robustness. A pioneer work claims that the texture bias of neural networks is key to model performance under image corruptions~\cite{geirhos2018imagenet}. An empirical study indicates that increasing shape bias has no direct relationship with improving corruption robustness~\cite{mummadi2021does}. On the other hand, corruption robustness is analyzed from the Fourier perspective, which disentangles the effect of different image corruptions on neural networks in the low-and high-frequency domains~\cite {yin2019fourier}. In~\cite{yang2020boundary}, the authors connect the decision boundary thickness of neural networks with model robustness. Lately, a large-scale study shows that several assumptions of model bias do not hold from the perspective of network robustness~\cite{gavrikov2024can}. More relevant are works that implicitly model non-robust features induced by patch-based transformations in Vision Transformers, demonstrating that the model exploits these patterns in its predictions~\citep{qin2022understanding,guo2023improving}. To date, there has been no effort to explicitly model robust or non-robust features to better understand the corruption robustness.

\begin{figure*}[!ht]
    \centering
    \includegraphics[width=1.0\linewidth]{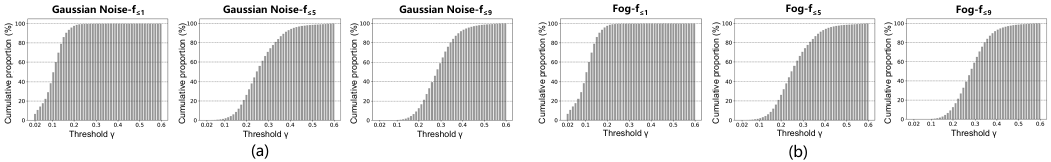}
    \caption{The cumulative distribution of $\gamma$-robust features at different sub-networks under the corruption of (a)~Gaussian Noise and (b)~Fog. The consistent shift toward higher thresholds in deeper layers indicates a progressive scarcity of robust features as the network depth increases, revealing that high-level representations are more sensitive to structured corruptions.}
    \label{Figure 2}
\end{figure*}
%-------------------------------------------------------------------------
\begin{figure}[!ht]
    \centering
    \includegraphics[width=1.0\linewidth]{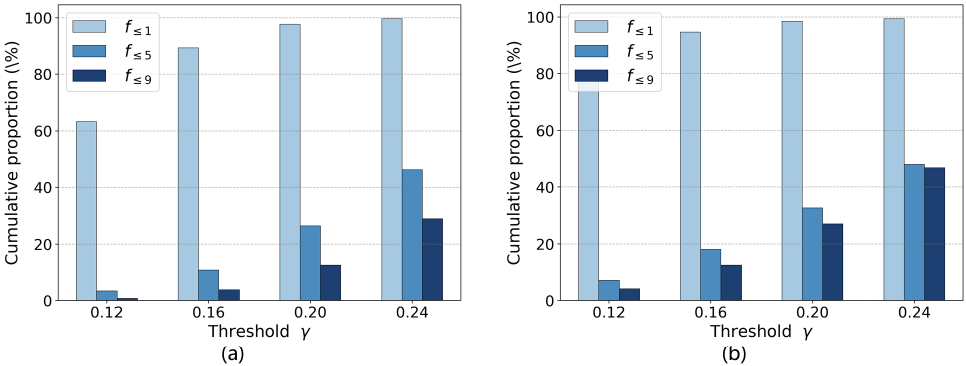}
    \caption{Robust feature proportions across thresholds $\gamma$ under (a) Gaussian Noise and (b) Fog. The downward trend across stages ($f_{\leq 1} \to f_{\leq 9}$) reveals systemic robustness decay.}
    \label{Figure 2_2}
\end{figure}
%-------------------------------------------------------------------------
\subsection{Enhancing Corruption Robustness}
Enhancing corruption robustness is a major research focus, with benchmarks spanning multiple vision tasks~\citep{hendrycks2019benchmarking,michaelis2019benchmarking,kamann2021benchmarking}. A dominant direction is data augmentation, which consistently improves robustness~\citep{geirhos2018imagenet,yin2019fourier,guo2023improving}. AugMix~\cite{hendrycks2019augmix} mixes diverse image operations, and later methods extend mixing to style~\citep{hendrycks2021many,zhou2024mixstyle}, spatial~\cite{modas2022prime}, and frequency domains~\citep{yucel2023hybridaugment++,vaish2024fourier}. Other lines combine augmentations via ensembles~\citep{saikia2021improving,diffenderfer2021winning} or propose training recipes coupling augmentation with regularization~\citep{wightman2021resnet,touvron2021training}. Beyond training pipelines, bio-inspired architectures and learning paradigms are explored~\citep{dapello2020simulating,teti2022lcanets,li2019learning,safarani2021towards}. Fine-tuning strategies also target corruption-robust sub-networks~\citep{lee2022surgical,guo2022improving}; among them, EWS~\cite{guo2022improving}, closely related to our approach, uses knowledge distillation with a model-specific controller to refine non-robust sub-networks. In contrast, our method requires no additional model structures and can be seamlessly integrated into existing deep-learning frameworks.

\section{Methodology}
\label{sec:methodology}

%----------------------------------------------------
\subsection{Problem Formulation}
We consider the robustness of a neural network $f_\theta$ under natural image corruptions. Formally, let $\mathcal{D}=\{(x_i,y_i)\}^{N}_{i=1}$ be a clean dataset from $\mathbb{P}(X,Y)$. A corruption function $h\in\mathcal{H}$ transforms $x_i$ into a corrupted version $h(x_i)$. Following the covariate shift assumption~\cite{gavrikov2024can}, where $\mathbb{P}(X) \neq \mathbb{P}(h(X))$ but the label remains invariant, our objective is to minimize the expected risk across the corruption space $\mathcal{H}$:
%%%%%%%%%%%%%%%%%%%%%%%%%%%%%%%%%
\begin{equation}
    \min_{\theta}\mathbb{E}_{h\sim H}[\mathbb{E}_{X,Y\sim \mathbb{P}(X,Y)}[\mathcal{L}(f_\theta(h(X)),Y)]],
\end{equation}\label{objective}
%%%%%%%%%%%%%%%%%%%%%%%%%%%%%%%%%
where $\mathcal{L}$ denotes a standard supervised loss (e.g., cross-entropy). Unlike adversarial robustness works that emphasize input-level sensitivity~\cite{ilyas2019adversarial}, we posit that for natural corruptions, robustness is fundamentally tied to the stability of internal features. To formalize this, we decompose $f_\theta$ into $L$ layers: $f_\theta(\cdot)= f_L \circ f_{L-1} \circ \cdots \circ f_1$, where $f_{\leq l}$ maps the input into a feature map $Z_l$. Each neuron in layer $l$ constitutes a \textbf{computational pathway} $P_{l,k}: X \to Z_{l,k}$~(abbreviated as $P$). The robustness of $f_\theta$ thus depends on whether these pathways can produce invariant features $Z_{l,k}$ under corruption $h$.

\subsection{Robust and Non-robust Features}
%---------------------------------------------------------------
\subsubsection{Characterizing robust features and pathways}
The concept of robust features was introduced by \cite{ilyas2019adversarial} via a pixel-level disentanglement framework to interpret adversarial examples. Subsequent studies extended this analysis to the layer~\cite{yan2021cifs} and neuron levels~\citep{zhang2020interpreting,madaan2020adversarial} to enhance adversarial robustness. However, while adversarial perturbations are typically high-dimensional, non-semantic noise affecting the local pixel manifold, natural corruptions (e.g., fog, blur) exhibit global structural patterns that manifest as semantic shifts within the latent space. Despite this, the nature of robust and non-robust features under such structured distortions remains poorly understood, particularly regarding their explicit disentanglement. To fill this gap, we formalize robust features under image corruptions through the lens of computational pathways.

\begin{definition}[Robust features]\label{def_robust_features}
Given a normalized distortion metric $D$, a feature $z_{l,k}$ is said to be $\gamma$-robust ($\gamma>0$) for corruption $h$ if
\begin{equation}
    D(z_{l,k},\hat{z}_{l,k}) < \gamma,
    \label{eq:robust_features}
\end{equation}
\end{definition}
where $\hat{z}_{l,k}$ denotes the corrupted version of $z_{l,k}$ induced by applying the corruption $h$ to the input.
%\textbf{Definition 3.1}~(Robust features). \textit{Given a normalized distortion metric $D$, a feature $z_{l,k}$ is said to be $\gamma$-robust ($\gamma>0$) for corruption $h$ if}
%\begin{equation}
%    D(z_{l,k},\hat{z}_{l,k})<\gamma
%\end{equation}
\begin{definition}[Robust pathways]\label{def_robust_pathways}
Given a normalized distortion metric $D$, a computational pathway $P$ is said to be $\gamma$-robust ($\gamma>0$) for corruption $h$ if
\begin{equation}
    \mathbb{E}_{x\sim \mathbb{P}(X)}[D(P(x), P(h(x)))] < \gamma.
    %\mathbb{E}_{x\sim X}[D(z_{l,k},\hat{z}_{l,k})]< \gamma.
\end{equation}
\end{definition}
Note that these definitions isolate robustness as a property of features and pathways, independent of their impact on predictions; \cref{subsec: a_closer_look} later investigates this connection.
%%%%%%%%%%%%%%%%%%%%%%%%%%%%%%%%%
\begin{figure*}[t]
    \centering
    \includegraphics[width=1.0\linewidth]{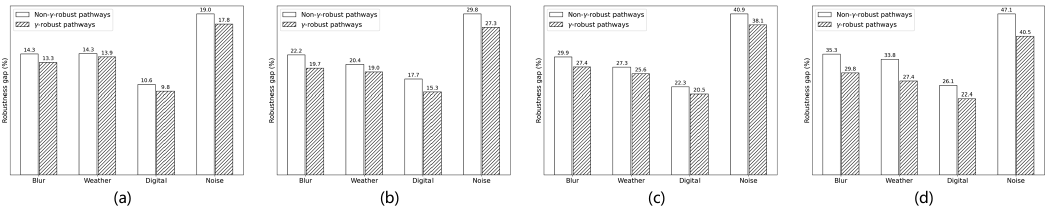}
    \caption{Impact of pathway robustness on sub-networks and the full model. We report the robustness gap (lower is better) for (a) $f_{\leq 1}$, (b) $f_{\leq 5}$, (c) $f_{\leq 9}$, and (d) $f_{\theta}$ when preserving either $\gamma$-robust or non-$\gamma$-robust pathways.}
    \label{Figure 3}
\end{figure*}
%%%%%%%%%%%%%%%%%%%%%%%%%%%%%%%%%
\begin{figure}[t]
    \centering
    \includegraphics[width=1.0\linewidth]{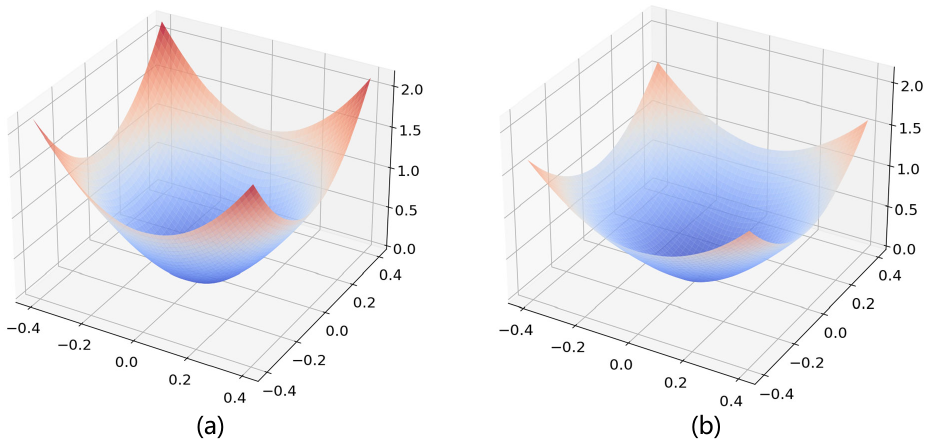}
    \caption{Loss landscapes of $f_\theta$ under input perturbations. (a) Non-$\gamma$-robust pathways exhibit a sharp landscape, whereas (b) $\gamma$-robust pathways show a flatter surface.}
    \label{Figure 3_2}
\end{figure}

%--------------------------------------------------------------
\subsubsection{Identifying robust features: Layer-wise distribution and evolution}
To demystify the internal behavior of robust features, we quantify their layer-wise distribution across the network. We employ the local Lipschitz constant \citep{cisse2017parseval,szegedy2013intriguing} as the distortion metric $D$ in \cref{eq:robust_features}, denoted as $D_{\text{Lip}}$:
\begin{equation}
D_{\text{Lip}}(z_{l,k},\hat{z}_{l,k})=\frac{||z_{l,k}-\hat{z}_{l,k}||}{||x-h(x)||},
\label{lipschitz}
\end{equation}
where $||\cdot||$ represents the normalized Frobenius norm. This metric effectively captures the sensitivity of internal features relative to input-level perturbations. Using \cref{eq:robust_features}, \Cref{Figure 2}~(a)-(b) visualize the cumulative distribution of $\gamma$-robust features for a pretrained ResNet18 model under Gaussian noise and Fog, respectively. To provide a more intuitive and quantitative comparison, \Cref{Figure 2_2} summarizes the proportion of robust features at specific thresholds $\gamma$.

Our empirical analysis reveals a non-homogeneous evolution of robustness along the network depth. Specifically, while shallow sub-networks (e.g., $f_{\leq 1}$) retain a high proportion of $\gamma$-robust features, this proportion exhibits a progressive decay as information propagates into deeper sub-networks (e.g., $f_{\leq 5}$ and $f_{\leq 9}$). As evidenced by the sharp contrast in \Cref{Figure 2_2}, the available pool of robust features shrinks drastically in deeper layers, particularly at stringent thresholds. This trend suggests that as receptive fields expand to capture global structural patterns, features become increasingly susceptible to semantic shifts induced by structured corruptions. Consistent results for additional corruption types are further detailed in Appendix~\ref{subsec: extended visualization}, underscoring that internal robustness degradation is a pervasive challenge across diverse distortion categories.
\begin{remark}\label{remark_1}
Under image corruptions, robust features exhibit a progressive decay across neural layers, indicating that their prevalence diminishes significantly as features propagate into deeper stages of the network.
\end{remark}

\begin{figure*}[!ht]
    \centering
    \includegraphics[width=0.95\linewidth]{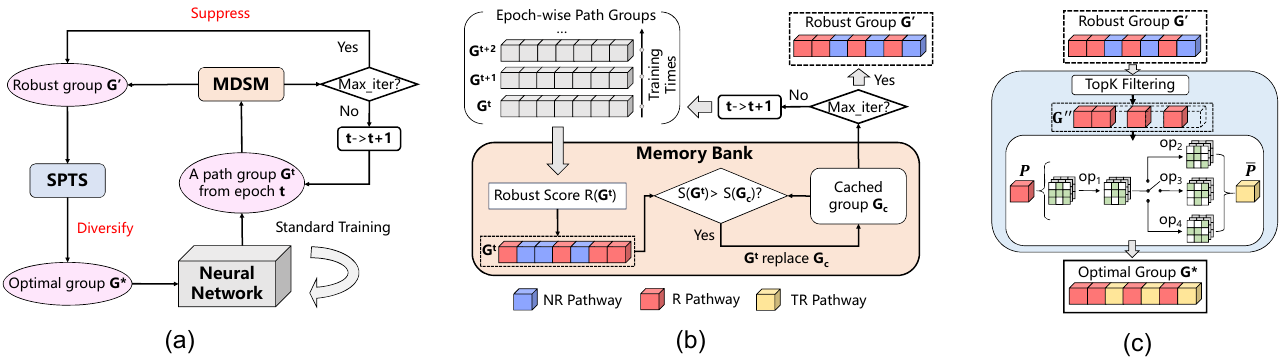}
    \caption{The pipeline of (a)~S\&D, (b)~MDSM, and (c)~SPTS. NR, R and TR denote non-robust, robust and tweaked robust pathways, respectively. }
    \label{Figure 4}
\end{figure*}

\subsubsection{A closer look at robust features: causal impact and loss landscapes}
\label{subsec: a_closer_look}
Based on the above observation, we take a closer look at the collective behavior of robust~(non-robust) features under image corruptions. To explore this, we investigate the causal role of these features by isolating robust and non-robust pathways and quantifying their impact on the robustness of both downstream sub-networks and the full model. For a given corruption $h$, we score each computational pathway $P$ in the stem stage ($l_1$) of a pretrained ResNet-18 using the empirical mean of $D_{\text{Lip}}$:
%%%%%%%%%%%%%%%%%%%%%%%%%%%%%%%%%
\begin{equation}
\bar{D}_{\text{Lip}}(P, h) = \frac{1}{N}\sum_{i=1}^{N} \frac{||P(x_i)-P(h(x_i))||}{||x_i-h(x_i)||}.
%\frac{1}{N}\sum_{i=1}^{N}D_{\text{Lip}}(z_{l_1,k}^i,\hat{z}_{l_1,k}^i),
\label{eq: robust_pathways}
\end{equation}
%%%%%%%%%%%%%%%%%%%%%%%%%%%%%%%%%
To evaluate their causal role, we partition pathways into $\gamma$-robust and non-$\gamma$-robust groups based on the top and bottom 50\% of scores. We selectively activate one group at layer $l_1$ while zeroing the other. For a fair comparison, we perform linear probing by attaching and fine-tuning a linear head on the final feature maps of each sub-network $f_{\leq l_2}$ and the full model $f_\theta$ using clean data. Comprehensive analysis across 15 corruption types in Appendix~\ref{subsec: fine-grained analysis} further confirms the consistent advantage of $\gamma$-robust pathways.

We evaluate the influence of these pathway groups using the robustness gap~\citep{taori2020measuring,fang2022data,tu2025toward}, defined as the accuracy degradation from clean to corrupted data. \Cref{Figure 3} reveals a consistent trend across all sub-networks and the full model: $\gamma$-robust pathways yield a significantly narrower gap compared to their non-robust counterparts. This disparity suggests that model robustness is sustained by this specific subset of invariant features. To provide a geometric explanation, we visualize the loss landscapes of $f_\theta$ in \Cref{Figure 3_2}. Notably, $\gamma$-robust pathways exhibit a significantly flatter loss surface, which ensures stable predictions in the vicinity of clean inputs and effectively minimizes the robustness gap.

%%%%%%%%%%%%%%%%%%%%%%%%%%%%%%%%%
\begin{remark}\label{remark_2}
The robustness of the full model is functionally dependent on the prevalence of robust features, which exert a direct impact on preserving performance across downstream layers and sub-networks under corruptions.
\end{remark}
%%%%%%%%%%%%%%%%%%%%%%%%%%%%%%%%%

%--------------------------------------------------------------------------------
\subsection{The S\&D Refinement}
%--------------------------------------------------------------------------------
\subsubsection{Overview}
Motivated by the hierarchical decay of robust features, we propose the S\&D refinement to enhance corruption robustness through pathway manipulation (\cref{Figure 4}~(a)). S\&D consists of two synergistic components: (1) MDSM, which identifies a stable pathway group $G'$ generating robust representations, and (2) SPTS, which constructs an optimal group $G^*$ by diversifying $G'$ while preserving its structural invariants. Notably, S\&D is non-intrusive, requiring no architectural modifications or interference with standard training, ensuring seamless integration into mainstream frameworks.

\subsubsection{MDSM: Selecting robust pathways}
The Memory-aware Dynamic Selection Mechanism (MDSM) identifies robust pathways by maintaining a candidate group $G_c$ in a memory bank (\cref{Figure 4}~(b)). At epoch $t$, a candidate collection 
$G^t = \{P_{l,1}, \dots, P_{l,K}\}$ is deterministically constructed from the current model state, with fixed indexing inherited from layer $l$. This collection is evaluated against $G_c$ using a fitness score $\mathcal{S}$ that balances feature stability with discriminability:
%%%%%%%%%%%%%%%%%%%%%%%%%%%%%%%%%
\begin{equation} \mathcal{S}(G^t) = \frac{1}{K}\sum_{k=1}^{K} [R(P_{l,k}(x), P_{l,k}(\hat{x}))] + \lambda \cdot \sigma(n/N), \label{eq: R_group} \end{equation}
where $X$ and $\hat{X}$ represent a batch of clean images and their corresponding corrupted versions. $R$ is a similarity metric (e.g., $L_2$ or CKA), and $n$ and $N$ denote the current and total scheduled MDSM steps at the epoch level. $\lambda$ is a hyperparameter balancing two terms and is set to 1. The second term, governed by a sigmoid-like schedule $\sigma(\cdot)$, prioritizes discriminability as training matures, inspired by findings that deeper layers develop more discriminative representations over time~\citep{zeiler2014visualizing,yosinski2014transferable,zhang2021understanding}. Using \cref{eq: R_group}, the memory bank iteratively updates $G_c$ over $N$ steps to output the robust pathway group $G'$. Ablations in Appendix~\ref{subsec: choice of similarity} and~\ref{subsec: choice of transforms} confirm that S\&D yields consistent gains regardless of the specific choice of $R$ or the transformations for $\hat{x}$. Notably, MDSM operates during epoch gaps, ensuring zero interference with standard training gradients.

\begin{table*}[ht]
    \centering
    \caption{We report Top-1 Accuracy~($\uparrow$) for ImageNet and mean Corruption Error (mCE) for its corrupted variants (C, $\bar{\text{C}}$, 3DCC, and V2-C). Avg.mCE~($\downarrow$) averages these four error rates. $[\cdot]$ indicates ensemble-based results; $(+)$ and $(-)$ denote performance changes via S\&D. $^*$ indicates the use of modern training recipes, including stronger augmentation and optimization practices.}
    \small
    \resizebox{0.88\textwidth}{!}{
    \begin{tabular}{ccccccccc}
    \textbf{Main} & \bf S\&D  & \bf ImageNet &\bf ImageNet-C & \bf ImageNet-$\bar{C}
    $ &\bf ImageNet-3DCC &\bf ImageNetV2-C & Avg.mCE~($\downarrow$) \\
    \hline
    \multicolumn{8}{c}{Backbones of Neural Network}\\
    %\multirow{2}{*}{SqueezeNet1\_1} & \ding{55}  & 57.3 & 18.0 & 24.7 & 28.0 & 13.0 & 20.9 \\
    %  &\cellcolor{gray!20} \ding{52}  & \cellcolor{gray!20} 58.3~[57.9] &\cellcolor{gray!20} 19.6 &\cellcolor{gray!20} 25.3 &\cellcolor{gray!20} 28.9 &\cellcolor{gray!20} 14.2 &\cellcolor{gray!20}22.0~(\textbf{+1.1}) \\
     \multirow{2}{*}{MobileNetV2~\cite{sandler2018mobilenetv2}} & \ding{55}  & 71.4 & 86.2 & 86.4 & 82.8 & 89.8 & 86.3\\
      &\cellcolor{gray!20} \ding{52}  & \cellcolor{gray!20} 71.6~[71.2] &\cellcolor{gray!20} 84.9 &\cellcolor{gray!20} 86.6 &\cellcolor{gray!20} 82.6 &\cellcolor{gray!20} 89.2 &\cellcolor{gray!20}85.8~(\textbf{-0.5})  \\
     \multirow{2}{*}{ResNet-50~\cite{he2016deep}} & \ding{55}  & 75.7 & 76.7 & 79.4 & 73.2 & 83.1 & 78.1\\
      &\cellcolor{gray!20} \ding{52}  & \cellcolor{gray!20} 74.5~[75.6] &\cellcolor{gray!20}72.4  &\cellcolor{gray!20}78.2  &\cellcolor{gray!20}71.3  &\cellcolor{gray!20}79.2  &\cellcolor{gray!20} 75.3~(\textbf{-2.8})\\
     \multirow{2}{*}{VGG19~\cite{simonyan2014very}} & \ding{55}  & 73.9 & 81.6 & 84.6 & 78.6 & 86.4 & 82.8 \\
      &\cellcolor{gray!20} \ding{52}  & \cellcolor{gray!20} 75.9~[74.2] &\cellcolor{gray!20}79.7  &\cellcolor{gray!20}81.1  &\cellcolor{gray!20}76.9  &\cellcolor{gray!20}84.9  &\cellcolor{gray!20} 80.7~(\textbf{-2.1}) \\
     \multirow{2}{*}{WideResNet-50\_2~\cite{zagoruyko2016wide}} & \ding{55}  & 78.3 & 71.7 & 74.4 & 68.9 & 79.2 & 73.6\\
      &\cellcolor{gray!20} \ding{52}  & \cellcolor{gray!20} 76.3~[78.3] &\cellcolor{gray!20} 68.5  &\cellcolor{gray!20} 73.9 &\cellcolor{gray!20} 67.7 &\cellcolor{gray!20} 76.1 &\cellcolor{gray!20}71.6~(\textbf{-2.0})\\
    \hline
    \multicolumn{8}{c}{Data Augmentation \& Model Regularization}\\
     \multirow{2}{*}{AugMix~\cite{hendrycks2019augmix}} & \ding{55}  & 76.2 & 72.8 & 74.8 & 70.3 & 79.6 & 74.4\\
      &\cellcolor{gray!20} \ding{52}  & \cellcolor{gray!20} 74.9~[76.2] &\cellcolor{gray!20}68.1  &\cellcolor{gray!20}75.4 &\cellcolor{gray!20}68.3  &\cellcolor{gray!20}75.7  &\cellcolor{gray!20}71.9~(\textbf{-2.5}) \\
     \multirow{2}{*}{TriAug~\cite{muller2021trivialaugment}} & \ding{55}  & 76.7 & 70.3 & 75.5 & 67.5 & 77.6 & 72.7\\
      &\cellcolor{gray!20} \ding{52}  & \cellcolor{gray!20} 75.3~[76.7] &\cellcolor{gray!20}67.7  &\cellcolor{gray!20}75.7 &\cellcolor{gray!20}66.3  &\cellcolor{gray!20}75.4  & \cellcolor{gray!20}71.3~(\textbf{-1.4})  \\
     \multirow{2}{*}{Label Smoothing~\cite{muller2019does}} & \ding{55}  & 76.6 & 75.2 & 77.1 & 72.1 & 81.5 & 76.5 \\
      &\cellcolor{gray!20} \ding{52}  & \cellcolor{gray!20} 75.0~[76.6] &\cellcolor{gray!20}72.1  &\cellcolor{gray!20}77.3  &\cellcolor{gray!20}70.8  &\cellcolor{gray!20}79.0  &\cellcolor{gray!20}74.8~(\textbf{-1.7}) \\
     \multirow{2}{*}{Mixup~\cite{zhang2018mixup}} & \ding{55}  & 76.7 & 71.7 & 71.4 & 70.1 & 78.7 & 73.0 \\
      &\cellcolor{gray!20} \ding{52}  & \cellcolor{gray!20} 74.9~[76.7] &\cellcolor{gray!20}68.1  &\cellcolor{gray!20}71.1  &\cellcolor{gray!20}68.8  &\cellcolor{gray!20}75.6  &\cellcolor{gray!20}70.9~(\textbf{-2.1}) \\
    \hline
    \multicolumn{8}{c}{Advanced Training Recipe}\\
     \multirow{2}{*}{MobileNetV2$^*$~\cite{sandler2018mobilenetv2}} & \ding{55}  & 72.1 & 86.1 & 82.0 & 81.3 & 89.2 &84.7 \\
      &\cellcolor{gray!20} \ding{52}  & \cellcolor{gray!20}  72.5~[72.3]
    &\cellcolor{gray!20}78.5  &\cellcolor{gray!20}78.3 &\cellcolor{gray!20}76.3  &\cellcolor{gray!20}83.7  &\cellcolor{gray!20} 79.2~(\textbf{-5.5})\\
     \multirow{2}{*}{ResNet-50$^*$~\cite{he2016deep}} & \ding{55}  & 80.7 & 66.5 & 66.1 & 62.3 & 73.7 &67.2 \\
      &\cellcolor{gray!20} \ding{52}  & \cellcolor{gray!20} 79.5~[79.7] 
    &\cellcolor{gray!20}61.5  &\cellcolor{gray!20}65.0  &\cellcolor{gray!20}60.9  &\cellcolor{gray!20}69.8  &\cellcolor{gray!20} 64.3~(\textbf{-2.9}) \\
     %\multirow{2}{*}{ConvNeXt-tiny$^*$} & \ding{55}  & 82.5 & 52.0 & 55.7 & 56.7 & 41.0 & 51.4\\
    %  &\cellcolor{gray!20} \ding{52}  & \cellcolor{gray!20} 81.8~[82.1] 
    %&\cellcolor{gray!20} 55.0 &\cellcolor{gray!20} 56.7 &\cellcolor{gray!20} 58.2 &\cellcolor{gray!20} 43.4  &\cellcolor{gray!20} 53.3~(\textbf{+1.9})\\
     \multirow{2}{*}{ConvNeXt-base$^*$~\cite{liu2022convnet}} & \ding{55}  & 84.0 & 53.6 & 53.1 & 53.1 & 63.5 & 55.8\\
      &\cellcolor{gray!20} \ding{52}  & \cellcolor{gray!20} 82.9~[83.4] 
    &\cellcolor{gray!20}50.7  &\cellcolor{gray!20}50.1  &\cellcolor{gray!20}51.8  &\cellcolor{gray!20}61.2  &\cellcolor{gray!20} 53.5~(\textbf{-2.3})  \\
    \hline
    \end{tabular}}
    \label{table 1}
\end{table*}
\begin{table}[!t]
    \centering
    \caption{Comparison with robust learning paradigms on ImageNet-C. We report Top-1 Accuracy~($\uparrow$). $^*$\ denotes methods that used either strong augmentation (S\&D$^*$) or improved training recipes ($\text{AdaSAP}_P^*$).}
    \small
    \resizebox{0.85\linewidth}{!}{
    %\fontsize{8}{10}\selectfont
    \begin{tabular}{cccc}
    \hline
    \bf Baseline  &\bf Stochastic Depth  & \bf $\text{AdaSAP}_P^*$ &\bf EWS \\
    39.2 & 38.9~(-0.3) & 43.3~(+4.1) & 40.6~(+1.4)   \\
    \hline
    \bf DST &\bf DAMP & \bf DAT  &\bf TVM \\
    38.7~(-0.5) & 41.4~(+2.2) & 41.1~(+1.9) & 39.8~(+0.6) \\
    \hline
     \bf VOneNet &\bf GaborNet & \bf S\&D &\bf $\text{S\&D}^*$ \\
     40.3~(+1.1) & 37.5~(-1.7) & 42.7~(+3.5) &  46.3~(+7.1) \\
    \hline
    \end{tabular}}
    \label{table 4}
\end{table}
%--------------------------------------------------------------------------
\subsubsection{SPTS: diversifying robust pathways}
To prevent overfitting to specific robust patterns, we propose the Structure-consistent Pathway Tweaking Strategy (SPTS) to diversify $G'$. As illustrated in \cref{Figure 4}~(c), SPTS first filters the top-$K$ most robust pathways:
%%%%%%%%%%%%%%%%%%%%%%%%%%%%%%%%%
\begin{equation}
G^{''} = \operatorname{TopK}(G^{'},R),
\label{topk}
\end{equation}
%%%%%%%%%%%%%%%%%%%%%%%%%%%%%%%%%
where $\operatorname{TopK}(\cdot)$denotes the operation that retains the $K$ pathways in $G^{'}$ according to their $R$ values, yielding the new group $G^{''}$. $K$ is typically set to $|G'|/2$; the impact of varying $K$ is detailed in Appendix~\ref{subsec: TopK}. To restore the group capacity, we generate augmented pathways $\bar{P}$ by applying symmetry-preserving transformations to the final weight matrix $A \in \mathbb{R}^{C \times W \times H}$ of pathways in $G''$:
%%%%%%%%%%%%%%%%%%%%%%%%%%%%%%%%%
\begin{align}
    &\bar{A}= \texttt{op}_i(\texttt{op}_1(A))\\
    \text{subject to} \quad &\texttt{op}_i\sim U\{\texttt{op}_2,\texttt{op}_3,\texttt{op}_4\},\notag
\end{align}
%%%%%%%%%%%%%%%%%%%%%%%%%%%%%%%%%
where $U$ denotes uniform sampling from a predefined set of operations. The transformation suite includes: \texttt{Channel shuffling} ($\mathtt{op}_1$), \texttt{Horizontal flipping} ($\mathtt{op}_2$), \texttt{Vertical flipping} ($\mathtt{op}_3$), and \texttt{Matrix transpose} ($\mathtt{op}_4$), with implementation details in Appendix~\ref{subsec: op}. This specific combination was identified as a highly effective configuration through comparative experiments in Appendix~\ref{subsec: choice of tweaking}. The augmented pathways $\bar{P}$ (parameterized by $\bar{A}$) introduce stochastic variations into the latent space while adhering to the robust hierarchical constraints of the original group. By merging $G''$ with these augmented pathways, we form the optimal group $G^*$. Finally, $G^*$ is integrated into the model $f_\theta$ for standard training, ensuring that the network is guided by a diverse yet stable set of computational pathways.

\section{Experiments}
\label{sec:experiments}

%----------------------------------
\subsection{Experimental Setup}
\textbf{Evaluation dataset.} 
The evaluation uses eight robustness benchmark datasets. For \textbf{image classification}, we use ImageNet-C~\cite{hendrycks2019benchmarking}, ImageNet-$\bar{C}$~\cite{mintun2021interaction}, and ImageNet-3DCC~\cite{kar20223d}. ImageNet-C includes 15 corruption types at five severity levels, ImageNet-$\bar{C}$ adds 10 new corruptions, and ImageNet-3DCC introduces 12 3D-based corruptions mimicking real-world distortions. Additionally, we create ImageNetV2-C by applying 15 corruption types to ImageNetV2~\cite{recht2019imagenet}. For \textbf{object detection}, we use COCO-C~\cite{michaelis2019benchmarking}, a corrupted version of COCO~\cite{lin2014microsoft}. For \textbf{semantic segmentation}, we generate VOC-C, Cityscapes-C, and ADE20K-C from PASCAL VOC 2012~\cite{everingham2015pascal}, Cityscapes~\cite{cordts2016cityscapes}, and ADE20K~\cite{zhou2019semantic}. All datasets are used exclusively for testing.

%We evaluate on eight robustness benchmark datasets. For \textbf{image classification}, we use ImageNet-C~\cite{hendrycks2019benchmarking}, ImageNet-$\bar{C}$~\cite{mintun2021interaction} and ImageNet-3DCC~\cite{kar20223d} for validation. ImageNet-C, a pioneering dataset, contains 15 corruption types at five severity levels. ImageNet-$\bar{C}$ introduces 30 new corruptions, while ImageNet-3DCC adds 20 3D-based corruptions that more closely mimic real-world distortions. We further construct ImageNetV2-C by applying 15 corruption types to ImageNetV2~\cite{recht2019imagenet} for more challenging scenarios. For \textbf{object detection}, we use COCO-C~\cite{michaelis2019benchmarking}, a corrupted version of COCO~\cite{lin2014microsoft}. For \textbf{semantic segmentation}, we generate VOC-C, Cityscapes-C, and ADE20K-C from test splits of PASCAL VOC 2012~\cite{everingham2015pascal}, Cityscapes~\cite{cordts2016cityscapes}, and ADE20K~\cite{zhou2019semantic}. All robustness datasets are used exclusively for testing.
\begin{table*}[t]
    \centering
    \caption{We report Top-1 Accuracy~($\uparrow$) for ImageNet-100 and mean Corruption Error (mCE) for its variants (C, $\bar{\text{C}}$, 3DCC, and V2-C). Avg. mCE~($\downarrow$) denotes the average error across these four benchmarks. $(+)$ and $(-)$ reflect the performance changes after applying S\&D.}
    \small
    \resizebox{0.7\textwidth}{!}{
    \begin{tabular}{ccccccccc}
    \hline
    \textbf{Main} & \bf S\&D  & \bf IN-100 &\bf IN-100-C & \bf IN-100-$\bar{C}$ &\bf IN-100-3DCC &\bf IN-100V2-C & Avg.mCE~($\downarrow$) \\
    \hline
    %\multirow{2}{*}{SqueezeNet1\_1} & \ding{55}  & 57.3 & 18.0 & 24.7 & 28.0 & 13.0 & 20.9 \\
    %  &\cellcolor{gray!20} \ding{52}  & \cellcolor{gray!20} 58.3~[57.9] &\cellcolor{gray!20} 19.6 &\cellcolor{gray!20} 25.3 &\cellcolor{gray!20} 28.9 &\cellcolor{gray!20} 14.2 &\cellcolor{gray!20}22.0~(\textbf{+1.1}) \\
     \multirow{2}{*}{MobileViT\_S} & \ding{55}  & 85.4 & 88.8 & 92.5 & 84.8 & 92.9 & 89.8  \\
      &\cellcolor{gray!20} \ding{52}  & \cellcolor{gray!20} 85.7  &\cellcolor{gray!20} 87.3 &\cellcolor{gray!20}91.6  &\cellcolor{gray!20}85.1  &\cellcolor{gray!20}91.6  &\cellcolor{gray!20}88.9~(\textbf{-0.9})  \\
     \multirow{2}{*}{EfficientFormer\_L1} & \ding{55}  & 91.6 & 73.7 & 66.3 & 65.2 & 80.4 & 71.4 \\
      &\cellcolor{gray!20} \ding{52}  & \cellcolor{gray!20} 91.8 &\cellcolor{gray!20}74.4  &\cellcolor{gray!20}63.4   &\cellcolor{gray!20}64.0  &\cellcolor{gray!20}80.6  &\cellcolor{gray!20} 70.6~(\textbf{-0.8}) \\
     \multirow{2}{*}{ViT\_Tiny} & \ding{55}  & 84.5 & 76.2 & 72.7 & 78.2 & 81.7 & 77.2 \\
      &\cellcolor{gray!20} \ding{52}  & \cellcolor{gray!20} 86.3  &\cellcolor{gray!20} 76.1 &\cellcolor{gray!20}70.5  &\cellcolor{gray!20}77.8  &\cellcolor{gray!20}80.3  &\cellcolor{gray!20} 76.2~(\textbf{-1.0}) \\
     \multirow{2}{*}{Mambaout\_femto} & \ding{55}  & 94.1 & 60.0 & 54.7 & 57.7 & 68.2 & 60.2 \\
      &\cellcolor{gray!20} \ding{52}  & \cellcolor{gray!20} 94.2  &\cellcolor{gray!20}57.6  &\cellcolor{gray!20}56.3   &\cellcolor{gray!20}57.5  &\cellcolor{gray!20}67.4  &\cellcolor{gray!20} 59.7~(\textbf{-0.5}) \\
    \hline
    \end{tabular}}
    \label{table s12}
\end{table*}

\begin{table*}[ht]
    \centering
    \caption{The semantic segmentation performance on ADE20K-C, Cityscapes-C and VOC-C. We report the mean Intersection over Union (mIoU $\uparrow$). \textcolor{gray}{\bf S} and \textcolor{gray}{\bf R} represent the results on clean and corrupted data, respectively.}
    \small
    \resizebox{0.7\textwidth}{!}{
    %\fontsize{8}{10}\selectfont
    \begin{tabular}{cccccccc}
    \hline
    \multirow{2}{*}{\bf Main} & \multirow{2}{*}{\textbf{S\&D}}  &\multicolumn{2}{c}{\bf ADE20K-C}  & \multicolumn{2}{c}{\bf Cityscapes-C} &\multicolumn{2}{c}{\bf VOC-C}\\
    & & \textcolor{gray}{\bf S} & \textcolor{gray}{\bf R} & \textcolor{gray}{\bf S} & \textcolor{gray}{\bf R} & \textcolor{gray}{\bf S} & \textcolor{gray}{\bf R}   \\
    \hline
    \multicolumn{8}{c}{Multi-scale Fusion-based Structure}\\
    \multirow{2}{*}{DeepLabV3+~\cite{chen2018encoder}} & \ding{55} & 42.1 & 21.3 & 78.6 & 36.2 & 75.9 & 44.9\\
     &\cellcolor{gray!25} \ding{52} &\cellcolor{gray!20} 41.2 &\cellcolor{gray!20} 23.1~(\textbf{+1.8}) &\cellcolor{gray!20} 77.9 &\cellcolor{gray!20}39.2~(\textbf{+3.0}) &\cellcolor{gray!20} 75.4 &\cellcolor{gray!20}48.0~(\textbf{+3.1})\\
    \multirow{2}{*}{PSPNet~\cite{zhao2017pyramid}} & \ding{55} & 41.2 & 20.1 & 77.2 & 34.3 & 75.9 & 44.5\\
     &\cellcolor{gray!20} \ding{52} &\cellcolor{gray!20} 40.0 &\cellcolor{gray!20}21.5~(\textbf{+1.4}) &\cellcolor{gray!20} 76.3 &\cellcolor{gray!20}37.8~(\textbf{+3.5}) &\cellcolor{gray!20} 76.0 &\cellcolor{gray!20}47.4~(\textbf{+2.9})\\
    \hline
    \multicolumn{8}{c}{Attention-based Structure}\\
    \multirow{2}{*}{Mask2Former~\cite{cheng2022masked}} & \ding{55} & 47.2 & 25.2 & 79.3 & 40.8 & -- & --\\
     &\cellcolor{gray!20} \ding{52} &\cellcolor{gray!20} 46.4 &\cellcolor{gray!20} 29.2~(\textbf{+4.0}) &\cellcolor{gray!20} 77.3 &\cellcolor{gray!20} 47.1~(\textbf{+6.3}) &\cellcolor{gray!20} -- &\cellcolor{gray!20} --\\
    \multirow{2}{*}{ANN~\cite{zhu2019asymmetric}} & \ding{55} & 39.6 & 19.2 & 76.5 & 33.6 & 74.9 & 43.3\\
     &\cellcolor{gray!20} \ding{52} &\cellcolor{gray!20} 38.9 &\cellcolor{gray!20}20.6~(\textbf{+1.4}) &\cellcolor{gray!20} 75.5 &\cellcolor{gray!20} 35.9~(\textbf{+2.3}) &\cellcolor{gray!20} 75.0 &\cellcolor{gray!20} 46.7~(\textbf{+3.4})\\
    \multirow{2}{*}{CCNet~\cite{huang2019ccnet}} & \ding{55} & 41.2 & 20.1 & 78.9 & 34.0 & 76.2 & 44.1\\
     &\cellcolor{gray!20} \ding{52} &\cellcolor{gray!20} 40.2 &\cellcolor{gray!20} 21.5~(\textbf{+1.4}) &\cellcolor{gray!20} 78.0 &\cellcolor{gray!20} 38.0~(\textbf{+4.0}) &\cellcolor{gray!20} 75.9 &\cellcolor{gray!20} 47.6~(\textbf{+3.5}) \\
    \multirow{2}{*}{DANet~\cite{fu2019dual}} & \ding{55} & 40.1 & 19.9 & 78.0 & 34.5 & 74.5 & 43.9\\
     &\cellcolor{gray!20} \ding{52} &\cellcolor{gray!20} 39.4 &\cellcolor{gray!20} 22.0~(\textbf{+2.1}) &\cellcolor{gray!20} 77.7 &\cellcolor{gray!20} 38.5~(\textbf{+4.0}) &\cellcolor{gray!20} 74.2 &\cellcolor{gray!20} 46.6~(\textbf{+2.7})\\
    \multirow{2}{*}{GCNet~\cite{cao2019gcnet}} & \ding{55} & 40.3 & 19.9 & 76.8 & 33.1  & 76.4 & 43.9\\
     &\cellcolor{gray!20} \ding{52} &\cellcolor{gray!20} 39.5 &\cellcolor{gray!20} 21.6~(\textbf{+1.7}) &\cellcolor{gray!20} 76.5 &\cellcolor{gray!20} 36.6~(\textbf{+3.5}) &\cellcolor{gray!20} 74.8 &\cellcolor{gray!20} 47.5~(\textbf{+3.6})\\
    \multirow{2}{*}{PSANet~\cite{zhao2018psanet}} & \ding{55} & 40.8 & 20.0 & 77.0 & 34.3  & 76.4 & 44.6\\
     &\cellcolor{gray!20} \ding{52} &\cellcolor{gray!20} 40.1 &\cellcolor{gray!20} 21.6~(\textbf{+1.6}) &\cellcolor{gray!20} 75.9 &\cellcolor{gray!20} 36.7~(\textbf{+2.4}) &\cellcolor{gray!20} 75.5 &\cellcolor{gray!20} 46.6~(\textbf{+2.0})\\
    \hline
    \end{tabular}}
    \label{table 2}
\end{table*}

\begin{table}[ht]
    \centering
    \caption{The object detection performance on COCO and COCO-C. We report the mean Intersection over Union (mAP $\uparrow$).}
    \small
    \resizebox{0.93\linewidth}{!}{
    %\fontsize{8}{10}\selectfont
    \begin{tabular}{cccc}
    \hline
    \bf Main  & \bf S\&D  &\bf COCO  & \bf COCO-C\\
    \hline
    \multicolumn{4}{c}{Two-Stage Detector}\\
     \multirow{2}{*}{Faster RCNN~\cite{ren2016faster}} & \ding{55} & 37.6 & 17.5\\
      &\cellcolor{gray!20} \ding{52} &\cellcolor{gray!20} 37.2 &\cellcolor{gray!20} 18.2~(\textbf{+0.7})\\
      \multirow{2}{*}{Mask RCNN~\cite{he2017mask}} & \ding{55} & 38.1 & 18.0\\
      &\cellcolor{gray!20} \ding{52} &\cellcolor{gray!20} 38.1 &\cellcolor{gray!20} 18.5~(\textbf{+0.5})\\
      \multirow{2}{*}{Cascade RCNN~\cite{cai2018cascade}} & \ding{55} & 40.5 & 18.9\\
      &\cellcolor{gray!20} \ding{52} &\cellcolor{gray!20} 40.5 &\cellcolor{gray!20} 19.5~(\textbf{+0.6})\\
      \multirow{2}{*}{Cascade Mask RCNN~\cite{cai2018cascade}} & \ding{55} & 41.1 & 19.3\\
      &\cellcolor{gray!20} \ding{52} &\cellcolor{gray!20} 41.1 &\cellcolor{gray!20} 19.9~(\textbf{+0.6})\\
    \hline
    \multicolumn{4}{c}{One-Stage Detector}\\
       \multirow{2}{*}{RetinaNet~\cite{ross2017focal}} & \ding{55} & 36.3 & 17.1\\
      &\cellcolor{gray!20} \ding{52} &\cellcolor{gray!20} 36.6 &\cellcolor{gray!20} 17.6~(\textbf{+0.5})\\ 
      \multirow{2}{*}{YOLOv5s~\cite{reis2023real}} & \ding{55} & 39.8 & 19.9\\
      &\cellcolor{gray!20} \ding{52} &\cellcolor{gray!20} 39.1 &\cellcolor{gray!20} 22.1~(\textbf{+2.2})\\ 
      \multirow{2}{*}{YOLOv8s~\cite{reis2023real}} & \ding{55} & 46.1 & 25.5\\
      &\cellcolor{gray!20} \ding{52} &\cellcolor{gray!20} 45.0 &\cellcolor{gray!20} 27.0~(\textbf{+1.5})\\ 
    \hline
    \end{tabular}}
    \label{table 3}
\end{table}
\textbf{Evaluation design.}
We compare S\&D with: (1) Sub-network selection (Stochastic Depth~\cite{huang2016deep}, AdaSAP~\cite{bairadaptive}); (2) Dynamic weight evolution (EWS~\cite{guo2022improving}, DST~\cite{wudynamic}, DAMP~\cite{trinh2024improving}); (3) Consistency learning (DAT~\cite{mao2022enhance}, TVM~\cite{saikia2021improving}); and (4) Bio-inspired design (VOneNet~\cite{dapello2020simulating}, GaborNet~\cite{perez2020gabor}). To demonstrate broad applicability, we evaluate S\&D across classification, detection, and segmentation using CNN and Transformer backbones, combined with data augmentations, regularization, and widely adopted training recipes (detailed in Appendix~\cref{table s2}). We further assess robustness under test-time adaptation and real-world corruptions. For fairness, all models follow standard protocols or their original settings. Metrics include clean accuracy/mAP/mIoU and mCE or average corrupted performance. Full baseline details are provided in Appendix~\ref{subsec: baseline methods}.

%--------------------------------------------------------------------
\textbf{Implementation details.}
S\&D is applied during the initial training phase ($N=5$ in \cref{eq: R_group}). We update the stem stage of each backbone as the updated pathway group $G^t$, a choice validated against deeper stages in Appendix~\ref{subsec: position}. We adopt CKA~\cite{kornblith2019similarity} as the default similarity metric $R$, which yields consistent gains across various metric alternatives (Appendix~\ref{subsec: choice of similarity}). We also explore an ensemble-based strategy for evaluating the clean images (Appendix~\ref{subsec: ensemble}). All classification models follow standard PyTorch recipes, while detection and segmentation tasks utilize ImageNet-pretrained ResNet-50 via MMDetection~\cite{chen2019mmdetection} and MMSegmentation~\cite{mmseg2020}. Ablation studies regarding transformation types and hyper-parameter sensitivity are detailed in Appendix~\ref{sec:analysis}.
\subsection{Experimental Results}

\subsubsection{Main results on robust classification}

\textbf{Comparison with baselines.}
We evaluate S\&D against diverse robust training paradigms: sub-network selection, weight evolution, consistency learning, and bio-inspired designs (\cref{table 4}). S\&D consistently achieves superior robust accuracy. Without strong augmentation, S\&D surpasses the second-best method, DAMP, by 1.3\%. When combined with strong augmentation ($^*$), S\&D yields a 7.1\% total gain over the baseline and outperforms $\text{AdaSAP}_P^*$ by 3.0\%. This confirms S\&D's effectiveness as a standalone strategy. For a fair comparison, results are compiled from original reports or our implementations; see Appendix~\ref{subsec: baseline methods} for details.

\textbf{Compatibility with training recipes.} \cref{table 1} shows S\&D’s efficacy across various backbones and configurations. We observe consistent gains of 0.4–2.5\% regardless of model scale. Notably, S\&D exhibits strong synergy with augmentations and regularizations, boosting robust accuracy by 2.2\% (AugMix) and 1.9\% (Mixup). Furthermore, under modern recipes, S\&D yields substantial gains of 4.3\% on MobileNetV2 and 2.0\% on ConvNeXt-Base. These results confirm S\&D as a versatile plug-and-play enhancement; see Appendix~\ref{subsec: results on image classification} for extended results.

\subsubsection{Versatility across architectures, tasks and scenarios}

\textbf{Architecture diversity.} We evaluate S\&D on Transformer and Mamba backbones using ImageNet-100 and its corruption variants (e.g., -C, -$\bar{C}$, -3DCC). S\&D consistently enhances robustness across these diverse architectures (\cref{table s12}). While gains are most pronounced for CNNs due to their texture bias~\cite{geirhos2018imagenet}, S\&D still yields steady improvements for ViT and Mamba models. These results demonstrate S\&D’s versatility across various architectural paradigms.

\textbf{Downstream tasks.}
To verify S\&D's versatility, we extend evaluation to dense prediction tasks. For semantic segmentation (\cref{table 2}), S\&D yields consistent mIoU gains across ADE20K-C (1.4--4.0\%), Cityscapes-C (2.3--6.3\%), and VOC-C (2.0--3.6\%), with Mask2Former achieving the largest boost. In object detection (\cref{table 3}), S\&D effectively enhances robust mAP for both two-stage (Faster R-CNN: 17.5$\to$18.2) and one-stage detectors (YOLOv5s: 19.9$\to$22.1) while maintaining competitive clean performance. These results confirm S\&D's broad applicability in enhancing robustness for diverse visual applications.

\begin{table}[!t]
    \centering
    \caption{The segmentation performance~(mIOU) on ACDC.}
    \small
    \resizebox{0.9\linewidth}{!}{
    %\fontsize{8}{10}\selectfont
    \begin{tabular}{ccccccc}
    \hline
    \bf Main  & \bf S\&D  &\bf Fog  & \bf Night &\bf Rain  & \bf Snow & \bf Avg.\\
    \hline
     \multirow{2}{*}{DeepLabV3+} & \ding{55} & 63.3  & 13.3  &  49.2& 46.0  & 43.0\\
      &\cellcolor{gray!20} \ding{52} &\cellcolor{gray!20} 64.1  &\cellcolor{gray!20} 11.0  &\cellcolor{gray!20} 49.5 &\cellcolor{gray!20} 47.8  &\cellcolor{gray!20} 43.1~(\textbf{+0.1}) \\
     \multirow{2}{*}{PSPNet} & \ding{55} & 62.0 & 9.7 & 45.2& 39.8 & 39.2 \\
      &\cellcolor{gray!20} \ding{52} &\cellcolor{gray!20} 62.8  &\cellcolor{gray!20} 10.9  &\cellcolor{gray!20} 46.7 &\cellcolor{gray!20} 42.7  &\cellcolor{gray!20} 40.8~(\textbf{+1.6})\\
     \multirow{2}{*}{Mask2Former} & \ding{55} & 63.0 & 25.5 & 49.8 & 50.0 & 47.1\\
      &\cellcolor{gray!20} \ding{52} &\cellcolor{gray!20} 64.0  &\cellcolor{gray!20} 23.7 &\cellcolor{gray!20} 48.3 &\cellcolor{gray!20} 51.4  &\cellcolor{gray!20} 46.9~(\textbf{-0.2})\\
     \multirow{2}{*}{CCNet} & \ding{55} & 61.4 & 8.7 & 49.9& 45.8 & 41.5 \\
      &\cellcolor{gray!20} \ding{52} &\cellcolor{gray!20} 63.6  &\cellcolor{gray!20}  13.9 &\cellcolor{gray!20} 44.5 &\cellcolor{gray!20} 42.9 &\cellcolor{gray!20} 41.2~(\textbf{-0.3})\\
     \multirow{2}{*}{DANet} & \ding{55} & 61.1 & 9.2 &46.3 &43.6  & 40.1\\
      &\cellcolor{gray!20} \ding{52} &\cellcolor{gray!20} 61.1 &\cellcolor{gray!20} 10.0  &\cellcolor{gray!20} 48.6 &\cellcolor{gray!20} 41.2  &\cellcolor{gray!20} 40.2~(\textbf{+0.1}) \\
     \multirow{2}{*}{GCNet} & \ding{55} & 57.7 & 5.6 & 45.1 & 36.8 & 36.3\\
      &\cellcolor{gray!20} \ding{52} &\cellcolor{gray!20} 61.5  &\cellcolor{gray!20} 10.7   &\cellcolor{gray!20} 45.6 &\cellcolor{gray!20} 42.7  &\cellcolor{gray!20} 40.1~(\textbf{+3.8})\\
    \hline
    \end{tabular}}
    \label{table 13}
\end{table}

\begin{table}[t]
    \centering
    \caption{Comparison of test-time adaptation Methods at the severity 5 on ImageNet-C (Without vs. With S\&D). We report Top-1 Accuracy~($\uparrow$).}
    \small
    \resizebox{0.9\linewidth}{!}{
    %\fontsize{8}{10}\selectfont
    \begin{tabular}{cccccc}
    \hline
    & \bf AdaContrast  & \bf Tent &\bf BN    & \bf EATA\\
    w/o S\&D & 34.9 & 37.3 & 31.5 & 42.1  \\
    w/ S\&D & 38.1~(+3.2) & 38.3~(+1.0) & 31.9~(+0.4) & 42.6~(+0.5)   \\
    \hline
    &\bf SANTA & \bf SAR  &\bf RMT &\bf ROTTA \\
     w/o S\&D  & 39.9 & 37.8 & 42.2 & 32.6 \\
      w/ S\&D  & 40.6~(+0.7) & 38.6~(+0.8) & 44.2~(+2.0) & 34.0~(+1.4) \\
    \hline
    \end{tabular}}
    \label{table 5}
\end{table}
\begin{figure}[t]
    \centering
    \includegraphics[width=1.0\linewidth]{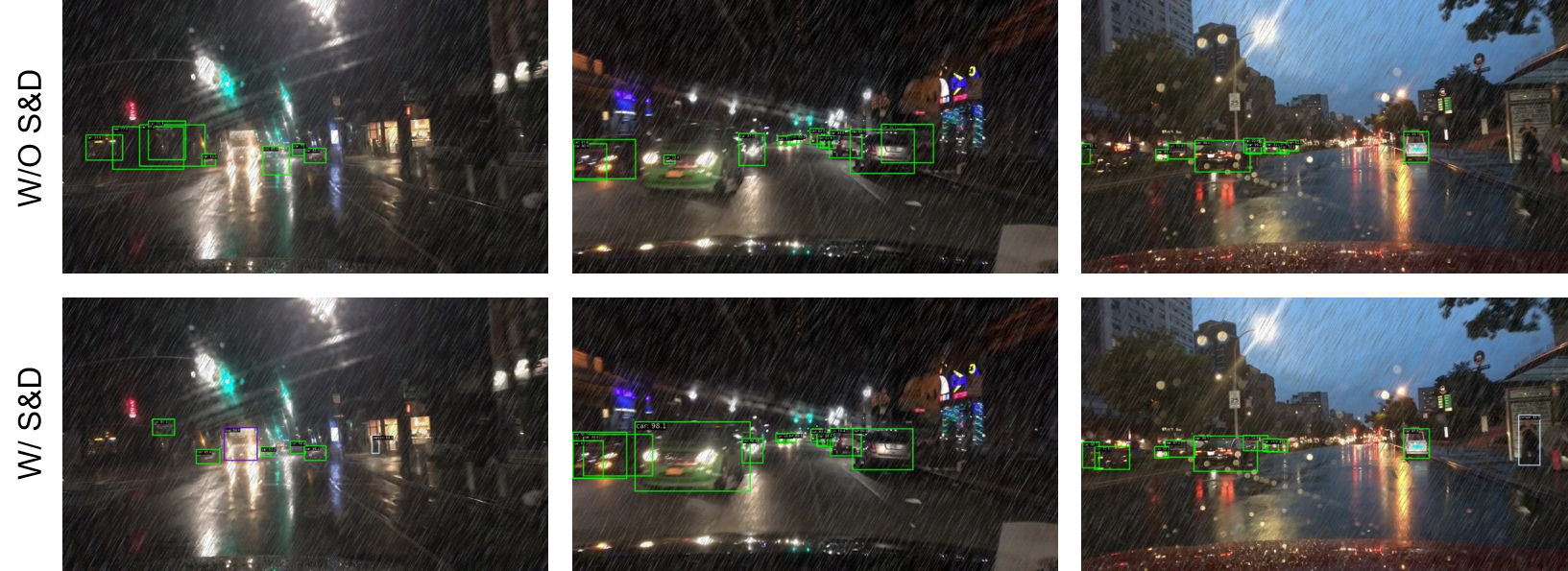}
    \caption{The detection results in night-rainy scenarios.}
    \label{Figure 6}
\end{figure}

\textbf{Real-world scenarios.} We first validate S\&D under real-world adverse weather for semantic segmentation and object detection. On the ACDC dataset~\cite{sakaridis2021acdc}, S\&D improves mIoU across most backbones (\cref{table 13}), notably boosting GCNet by 3.8\%, with corresponding segmentation visualizations provided in Appendix~\ref{subsec: semantic segmentation}. On the DWD benchmark~\cite{wu2022single}, S\&D enhances Faster R-CNN by 1.6 and 0.4 mAP in night-rainy and dusk-rainy scenarios, respectively. Qualitative results (\cref{Figure 6}) confirm S\&D’s efficacy in localizing and identifying objects under such compound corruptions. To further assess S\&D in dynamic environments, we evaluate continual test-time adaptation (TTA) on ImageNet-C. By integrating S\&D with eight representative methods, including Tent~\cite{wangtent}, BN~\cite{benz2021revisiting}, AdaContrast~\cite{chen2022contrastive}, EATA~\cite{niu2022efficient}, SANTA~\cite{chakrabarty2023santa}, SAR~\cite{niu2023towards}, RMT~\cite{dobler2023robust}, and ROTTA~\cite{yuan2023robust}, we observe consistent performance gains (\cref{table 5}) with a 3.2\% peak improvement. These results demonstrate S\&D’s broad compatibility in complex, real-world degraded environments.

\section{Analysis}
\label{sec:analysis_and_evaluation}

\begin{figure}[t]
    \centering
    \includegraphics[width=1.0\linewidth]{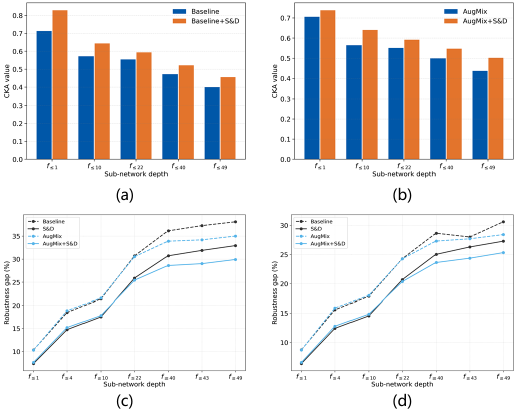}
    \caption{Comparison of (a) CKA between the baseline and S\&D, (b) CKA between AugMix and AugMix+S\&D, and robustness gap with and without S\&D on (c) ImageNet-C and (d) ImageNet-3DCC.}
    \label{Figure 5}
\end{figure}

\subsection{Representation Stability}
To evaluate representation stability, we compute CKA between ResNet-50 sub-network features ($f_{\leq1}$ to $f_{\leq49}$) under clean and corrupted inputs. As shown in \cref{Figure 5}~(a, b), S\&D consistently increases feature similarity across all depths. Further linear probing on ImageNet-C/3DCC using the robustness gap (lower is better) reveals consistent gap reductions in \cref{Figure 5}~(c, d). These findings confirm that S\&D effectively guides the network toward robust hierarchical representations in a bottom-up manner.

\begin{table*}[t]
\centering
\caption{Wall-clock overhead analysis of our method. ``Baseline'' denotes the runtime of the original training per epoch. ``MDSM'' denotes the additional per-epoch overhead introduced by our method, and ``SPTS'' is a one-time operation. Within MDSM, ``Aug'' denotes corrupted input generation, ``Fwd'' denotes forward propagation, and ``CKA'' denotes CKA computation. Relative overhead is measured with respect to the baseline epoch time.}
\resizebox{0.98\linewidth}{!}{
\begin{tabular}{lccccc}
\hline
\textbf{Component} & \textbf{Execution Frequency} & \textbf{Time (s)} & \textbf{Time Breakdown (s)} & \textbf{Against Baseline Epoch} \\
\hline
Baseline training & Every epoch & 2294.27 & -- & -- \\
MDSM & Selected epochs & 0.3233 $\pm$ 0.0015 & Aug: 0.0056,\; Fwd: 0.0420,\; CKA: 0.2757 & $\sim$0.01\% \\
SPTS & Last selected epoch & 0.0057 $\pm$ 0.0070 & -- & $<0.001\%$ \\
\hline
\end{tabular}}
\label{tab:overhead_rebuttal}
\end{table*}

\begin{table*}[t]
\centering
\caption{Throughput and memory overhead analysis of MDSM compared with baseline training. ``Peak Mem Diff'' is computed as MDSM peak memory minus baseline peak memory. Negative values indicate that MDSM remains within the baseline training memory budget.}
\resizebox{0.8\linewidth}{!}{
\begin{tabular}{ccccc}
\hline
\textbf{Baseline Throughput} & \textbf{MDSM Throughput} & \textbf{Baseline Peak Mem} & \textbf{MDSM Peak Mem} & \textbf{Peak Mem Diff} \\
\textbf{(img/s)} & \textbf{(img/s)} & \textbf{(MB)} & \textbf{(MB)} & \textbf{(MB)} \\
\hline
880.79 & 6003.23 & 21292.07 & 7640.79 & -13651.28 \\
\hline
\end{tabular}}
\label{tab:mdsm_profile}
\end{table*}

\subsection{Training Efficiency}
We measured the training-time and memory overhead of S\&D relative to a standard ResNet-50 baseline. \cref{tab:overhead_rebuttal} shows that baseline training requires 2294.27~s/epoch, while MDSM adds only 0.3233~s on selected epochs ($\sim$0.01\%), and the one-time SPTS cost is 0.0057~s ($<$0.001\%), with CKA computation dominating the small overhead. \cref{tab:mdsm_profile} reports peak GPU memory and throughput: MDSM consumes 7640.79~MB versus 21292.07~MB for baseline, while throughput increases due to lighter memory usage. Overall, S\&D introduces negligible wall-clock and memory overhead during training, confirming its practical efficiency without affecting test-time cost.

\section{Conclusion}
\label{sec:conclusion}
This paper investigates the phenomenon of robust features under image corruptions. Our analysis reveals that robust features exhibit a progressive decay across neural layers and establishes that model-wide robustness is functionally dependent on their prevalence. Building on these insights, we propose S\&D, a simple refinement that suppresses non-robust pathways and diversifies robust ones during training. S\&D is architecture-agnostic, parameter-free, and requires zero test-time overhead, enabling seamless plug-and-play integration. Extensive experiments demonstrate that S\&D significantly enhances corruption robustness and OOD generalization, offering a new perspective on building robust vision models.

Despite these improvements, several limitations remain. S\&D introduces a consistent clean–corruption trade-off, as pathways stabilized under corruptions may suppress features beneficial for clean accuracy. Evaluations have focused on moderate-scale models and datasets; scalability to larger models and full ImageNet-scale data remains to be fully explored. The current framework targets 2D/3D vision tasks, and applicability to other modalities is not yet established. Future work will focus on developing a theoretical framework for internal robust features, designing strategies to mitigate clean-accuracy degradation, and extending S\&D to additional modalities such as audio and multimodal tasks.

\section*{Impact Statement}
We propose Suppress \& Diversify (S\&D), a training-time refinement that enhances the robustness of deep neural networks under natural image corruptions. S\&D improves reliability in safety-critical vision systems such as autonomous driving, surveillance, and medical imaging, without adding inference overhead. It is architecture-agnostic and compatible with diverse CNN and Transformer backbones. Potential risks include over-reliance on automated decisions in untested scenarios; users should validate models across varied environments and combine with augmentation or uncertainty estimation. S\&D provides a principled approach to building reliable and generalizable visual models.

%%%%%%%%%%%%%%%%%%%%%%%%%%%%%%%%%%%%%%%%%%%%%%%%%%%%%%%%%%%%%%%%%%%%%%%%%%%%%%%

%%%%%%%%%%%%%%%%%%%%%%%%%%%%%%%%%%%%%%%%%%%%%%%%%%%%%%%%%%%%%%%%%%%%%%%%%%%%%%%

%%%%%%%%%%%%%%%%%%%%%%%%%%%%%%%%%%%%%%%%%%%%%%%%%%%%%%%%%%%%%%%%%%%%%%%%%%%%%%%


\bibliography{main}
\bibliographystyle{icml2026}

%%%%%%%%%%%%%%%%%%%%%%%%%%%%%%%%%%%%%%%%%%%%%%%%%%%%%%%%%%%%%%%%%%%%%%%%%%%%%%%
%%%%%%%%%%%%%%%%%%%%%%%%%%%%%%%%%%%%%%%%%%%%%%%%%%%%%%%%%%%%%%%%%%%%%%%%%%%%%%%
% APPENDIX
%%%%%%%%%%%%%%%%%%%%%%%%%%%%%%%%%%%%%%%%%%%%%%%%%%%%%%%%%%%%%%%%%%%%%%%%%%%%%%%
%%%%%%%%%%%%%%%%%%%%%%%%%%%%%%%%%%%%%%%%%%%%%%%%%%%%%%%%%%%%%%%%%%%%%%%%%%%%%%%
\clearpage
\appendix
\counterwithin{table}{section}
\counterwithin{figure}{section}
\counterwithin{equation}{section}
%\onecolumn
%\section{You \emph{can} have an appendix here.}

\section{Implementation Details}
\label{sec:impl}

\subsection{The Tweaking Operations}
\label{subsec: op}
The tweaking operations include four types, \texttt{Channel shuffling}, \texttt{Horizontal flipping}, \texttt{Vertical flipping}, \texttt{Matrix transpose}, abbreviated as \texttt{op1}, \texttt{op2}, \texttt{op3} and \texttt{op4}. Let $A\in\mathbb{R}^{C\times W\times H}$ be the weight matrix, and $c,w,h$ are the corresponding index, respectively. The tweaking operations can be formulated as:
\begin{align}
op_i(x) &= 
\begin{cases} 
A_{\sigma(c),w,h} & \text{if } i=1 \\
A_{c,W-w+1,h} & \text{if } i=2 \\
A_{c,w,H-h+1} & \text{if } i=3 \\
A_{c,h,w} & \text{if } i=4, \\
\end{cases}
\end{align}
where $\sigma$ is a random permutation that rearranges the C channels.

%%%%%%%%%%%%%%%%%%%%%%%%%%%%%%%%%%%%%%%%%%%%%%%%%%%%%%%%%%%%%%%%%%%%%%%%%%%%%%%
\subsection{Ensemble-based Strategy} 
\label{subsec: ensemble}
We additionally test a simple ensemble-based strategy on ImageNet. Specifically, for Model~A, we select the ImageNet-pretrained model provided by PyTorch, while Model~B uses the same architecture but incorporates S\&D, such as ResNet-50 and ResNet-50+S\&D. Inspired by~\cite{park2023understanding}, we find that the $l_2$ norm of the output at the first layer can serve as a confidence value to distinguish between non-corrupted and corrupted data. Given an unknown sample $x$, its corresponding feature is denoted as $z$. We select the threshold value as 0.5. The prediction $\hat{y}$ can be computed:
\begin{align}
\hat{y} &= 
\begin{cases} 
\text{Model A}(x) & \text{if  }||z||_2<0.5 \\
\text{Model B}(x) & \text{otherwise} 
\end{cases}
\end{align}
The results of Tab.~1 in the main text demonstrate that this ensemble-based approach is effective across most network architectures.
%%%%%%%%%%%%%%%%%%%%%%%%%%%%%%%%%%%%%%%%%%%%%%%%%%%%%%%%%%%%%%%%%%%%%%%%%%%%%%%
%%%%%%%%%%%%%%%%%%%%%%%%%%%%%%%%%%%%%%%%%%%%%%%%%%%%%%%%%%%%%%%%%%%%%%%%%%%%%%%
\subsection{Baseline Methods}
\label{subsec: baseline methods}
We integrate S\&D in representative methods to validate its effectiveness across image classification, object detection and semantic segmentation tasks. 

\textbf{Robustness-related paradigms.}We compare S\&D against four representative categories of robustness-enhancing methods: (1) Sub-network Selection, including AdaSAP~\cite{bairadaptive} (results cited from the original paper) and Stochastic Depth~\cite{huang2016deep} (reproduced following the original protocols); (2) Dynamic Weight Evolution, such as EWS~\cite{guo2022improving}, DST~\cite{wudynamic}, and DAMP~\cite{trinh2024improving}, where results are cited from their respective publications; (3) Consistency Learning, featuring DAT~\cite{mao2022enhance} (evaluated using the official checkpoint) and TVM~\cite{saikia2021improving} (reproduced following original settings); and (4) Bio-inspired Designs, such as VOneNet~\cite{dapello2020simulating} and GaborNet~\cite{perez2020gabor}, both of which are strictly reproduced.

\textbf{Image classification.} 
We evaluate four aspects: (1) neural network backbones, including convolutional architectures such as AlexNet~\cite{krizhevsky2012imagenet}, MobileNetV2~\cite{sandler2018mobilenetv2}, SqueezeNet~\cite{iandola2016squeezenet}, VGG~\cite{simonyan2014very}, ResNet~\cite{he2016deep}, ResNeXt~\cite{xie2017aggregated}, WideResNet~\cite{zagoruyko2016wide}, and ConvNeXt~\cite{liu2022convnet}, as well as attention- and state-space-based models including Vision Transformer~(ViT)~\cite{dosovitskiy2020image}, MobileViT~\cite{mehta2021mobilevit}, Efficientformer~\cite{li2022efficientformer}, and MambaOut~\cite{yu2025mambaout}; detailed results for ViT and Mamba-based architectures are reported in \cref{table s12}; (2) data augmentation~(e.g., AugMix~\cite{hendrycks2019augmix}, AutoAug~\cite{cubuk2018autoaugment}, RandAug~\cite{cubuk2020randaugment}, TriAug~\cite{muller2021trivialaugment}, Random Erasing~\cite{zhong2020random}); (3) model regularization~(e.g., CutMix~\cite{yun2019cutmix}, Mixup~\cite{zhang2018mixup}, and Label smoothing~\cite{muller2019does}) and (4) a modern Pytorch training recipe~\cite{paszke2019pytorch} combining diverse augmentations, regularization, and extended training schedules. 

\textbf{Object detection\&semantic segmentation.} 
We test both two-stage detectors~(e.g., Faster RCNN~\cite{ren2016faster}, Mask RCNN~\cite{he2017mask}, Cascade RCNN, and Cascade Mask RCNN~\cite{cai2018cascade})and one-stage detectors~(RetinaNet~\cite{ross2017focal}, YOLOv5 and YOLOv8~\cite{reis2023real}). For semantic segmentation, we evaluate on multi-scale fusion-based~(e.g., DeepLabV3+~\cite{chen2018encoder}, PSPNet~\cite{zhao2017pyramid}) and attention-based architectures~(e.g., Mask2Former~\cite{cheng2022masked}, ANN~\cite{zhu2019asymmetric}, CCNet~\cite{huang2019ccnet}, DANet~\cite{fu2019dual}, GCNet~\cite{cao2019gcnet}, and PSANet~\cite{zhao2018psanet}). 

%\textbf{Selective\&robust training.} 
% Specifically, under selective training, we consider sub-network regularization methods (e.g., Dropout~\cite{srivastava2014dropout}, Stochastic Depth~\cite{huang2016deep}) and sub-network selection approaches (e.g., Taylor~\cite{molchanov2016pruning}, FPGM~\cite{he2019filter}, $\text{AdaSAP}_P$~\cite{bairadaptive}). For robust training, we include methods such as Dynamic Sparse Training~\cite{wudynamic} and EWS~\cite{guo2022improving}.

\textbf{Test-time adaptation methods.} 
To further evaluate the compatibility and effectiveness of S\&D under distribution shifts, we integrate it with representative test-time adaptation (TTA) methods, including AdaContrast~\cite{chen2022contrastive}, BN statistics update~\cite{benz2021revisiting}, Tent~\cite{wangtent}, EATA~\cite{niu2022efficient}, SANTA~\cite{chakrabarty2023santa}, SAR~\cite{niu2023towards}, RMT~\cite{dobler2023robust} and RoTTA~\cite{yuan2023robust}, under the continual test-time adaptation setting. In this setting, models adapt sequentially to streaming corrupted data without access to the original training distribution. As shown in Tab.~5 of the main work, S\&D consistently improves performance across various TTA methods and under diverse corruption types.

\begin{table*}[ht]
    \centering
    \caption{Hyperparameters of the baseline training recipe and advanced training recipe.}
    %\fontsize{8}{10}\selectfont
    \resizebox{0.67\linewidth}{!}{
    \begin{tabular}{cccccc}
    \hline
      & \bf Baseline  &\bf Advanced & & \bf Baseline  &\bf Advanced \\
    \hline
     Train Res & 224 & 176 & Label smoothing $\epsilon$ & \textcolor{gray}{\ding{55}} & 0.1\\
     Test Res & 224 & 232 & Repeated Aug &\textcolor{gray}{\ding{55}}  & 4\\
     Epochs & 90 & 600 & H. flip & \ding{52}   &\ding{52} \\
     Batch size & 256 & 1024 & RRC & \ding{52}     & \ding{52}  \\
     Optimizer & SGD-M & SGD-M & Tri Augment &\textcolor{gray}{\ding{55}} & \ding{52}\\
     LR & 0.1 & 0.5 & Mixup alpha &\textcolor{gray}{\ding{55}} & 0.2\\
     LR Decay & step & cosine & Cutmix alpha &\textcolor{gray}{\ding{55}} & 1.0\\
     decay rate & 0.1 & - & Erasing prob. &\textcolor{gray}{\ding{55}} & 0.1\\
     decay epochs & 30 & - & ColorJitter &\ding{52}  & \ding{52} \\
     Weight decay & $10^{-4}$ & $2\times 10^{-5}$ & EMA &\textcolor{gray}{\ding{55}}  & \ding{52} \\
     Warmup Epochs & \textcolor{gray}{\ding{55}} & 5 & CEloss &\ding{52}  &\ding{52}\\
     \hline
    \end{tabular}}
    \label{table s2}
\end{table*}

\subsection{Training Configurations}

\textbf{Neural network backbone.} Convolutional architectures are trained using the official PyTorch training recipes~\cite{paszke2019pytorch}. For the Vision Transformer (ViT), we follow the training protocol of DeiT~\cite{touvron2021training}, trained from scratch without knowledge distillation. Other Transformer-based and Mamba-based models are trained from scratch according to the original procedures described in their respective papers, using implementations from the timm library~\cite{rw2019timm}. 

\textbf{Data augmentation and model regularization.} We consistently use ResNet-50 with the following configuration: 90 epochs, batch size of 256, initial learning rate of 0.1, SGD optimizer, and a StepLR scheduler that decays the learning rate by a factor of 0.1 every 30 epochs. All techniques are implemented using standard PyTorch modules, with hyperparameters listed in \cref{table s1}. 
\begin{table}[ht]
    \centering
    \caption{Hyperparameter settings for data augmentation and model regularization strategies.}
    \small
    \resizebox{0.8\linewidth}{!}{
    %\fontsize{8}{10}\selectfont
    \begin{tabular}{ccc}
    \hline
      \bf AugMix   &\bf RandAug  &\bf Random Erasing\\
      severity=3 & magnitude=9  & probability=0.1\\
    \hline
      \bf CutMix & \bf Mixup  &\bf Label Smoothing \\
       $\alpha$=0.1 & $\alpha$=0.1 &$\epsilon$=0.1 \\
     \hline
    \end{tabular}}
    \label{table s1}
\end{table}

\textbf{Advanced training recipe.} We adopt a new PyTorch training protocol for validation and conduct experiments on MobileNetV2, ResNet-50, ResNeXt-50 and ConvNeXt. The comparison with the baseline is shown in \cref{table s2}. 

\textbf{Test-time adaptation method.} We evaluate performance under continual test-time adaptation on ImageNet-C at severity level 5, using the same open-source codebase~\cite{dobler2024lost}, with the batch size set to 64 for all methods.

\section{Ablation Studies and Analysis}
\label{sec:analysis}

\subsection{Ablation of MDSM and SPTS}
We conduct ablation studies on the two core components of S\&D. As shown in \cref{table s3}, applying only MDSM slightly improves ImageNet-C robustness (33.5 vs. 32.9), while combining MDSM with SPTS yields the best corruption robustness (34.8), with minimal impact on clean accuracy. 
\begin{table}[h]
    \centering
    \caption{Ablation results of S\&D using ResNet-18 on ImageNet and ImageNet-C. Results are reported in top-1 accuracy (\%), with the highest values indicated in \textbf{bold}.}
    \small
    \resizebox{0.8\linewidth}{!}{
    %\fontsize{8}{10}\selectfont
    \begin{tabular}{cccc}
    \hline
     \bf MDSM & \bf SPTS & \bf ImageNet  &\bf ImageNet-C  \\
    \hline
      \textcolor{gray}{\ding{55}} & \textcolor{gray}{\ding{55}}& 69.2 & 32.9  \\
     \ding{52}&  \textcolor{gray}{\ding{55}}& \textbf{69.3} & 33.5  \\
     \ding{52}& \ding{52} & 68.6 &  \textbf{34.8} \\
     \hline
    \end{tabular}}
    \label{table s3}
\end{table}

\subsection{Choice of Tweaking Operations}
\label{subsec: choice of tweaking}
In SPTS, we adopt GeoDiversify, a geometric diversification strategy based on permutation and spatial transformations, to enhance robust feature diversity while preserving structural consistency. See \cref{subsec: op} for details. We also explore several other structure-consistent tweaking operations, including \textbf{WeightMix}, which performs weighted mixing between robust and non-robust pathways; \textbf{GaussPerturb}, which adds Gaussian noise to the weights of robust pathways; \textbf{FreqPerturb}, which applies perturbations in the frequency domain; \textbf{LowRankPerturb}, which applies perturbations along the smallest singular vector after SVD; and \textbf{UniPerturb}, which injects uniform noise into robust pathway weights. Results in \cref{table s8} show that \textbf{GeoDiversify} achieves the best trade-off between robustness gain and structural fidelity, and thus we use it as the default operation in our main experiments.
\begin{table}[h]
    \centering
    \caption{The results of different tweaking operations using ResNet-18 for ImageNet, ImageNet-C, and ImageNetV2-C. The highest accuracy is indicated in \textbf{bold}.}
    \resizebox{1.0\linewidth}{!}{
    %\fontsize{8}{10}\selectfont
    \begin{tabular}{cccc}
    \hline
      & \bf ImageNet  &\bf ImageNet-C &\bf ImageNetV2-C\\
    \hline
     Baseline & \bf 69.2 & 32.9 & 24.4\\
     WeightMix & 69.1 &  34.9 & 26.3 \\
     GaussPerturb & 68.6 & 35.3 & 26.6 \\
     UniPerturb & 68.5 & 35.5 & 26.7 \\
     FreqPerturb &  68.6 & 35.5 & \bf 26.8 \\
     LowRankPerturb & 68.4 & 35.3 & 26.6 \\
     GeoDiversify & 68.4 & \bf 35.6 & \bf 26.8 \\
     \hline
    \end{tabular}}
    \label{table s8}
\end{table}

\subsection{Choice of Corruption-generating Transforms}
\label{subsec: choice of transforms}
As mentioned in Sec.~3.3.2, we synthesize corrupted inputs $\hat{X}$ from clean images $X$ using simple transformations to guide robust pathway selection in MDSM. We evaluate several corruption strategies: \textbf{Contrast+Noise}~\cite{saikia2021improving}, combining contrast adjustment and additive noise; \textbf{ColorJitter}, randomly perturbing brightness, contrast, saturation, and hue; \textbf{LowFreq}, adding structured low-frequency perturbations; \textbf{MixedFreq}, injecting noise in both low- and high-frequency components; \textbf{Solarize}, randomly inverting pixels above a threshold; \textbf{AutoAugment}~(AutoAug), applying ImageNet-optimized policies; and \textbf{RandAugment}~(RandAug), using randomized augmentations with uniform magnitude. Results in \cref{table s9} show that while all methods provide useful supervision for MDSM, \textbf{Contrast+Noise} achieves the best trade-off between simplicity and effectiveness, and is thus adopted in our main experiments.

\begin{table}[h]
    \centering
    \caption{The results of ResNet-18 under different corruption-generating transforms on ImageNet, ImageNet-C, and ImageNetV2-C. Results are reported in top-1 accuracy (\%), with the highest values indicated in \textbf{bold}.}
    \resizebox{1.0\linewidth}{!}{
    %\fontsize{8}{10}\selectfont
    \begin{tabular}{cccc}
    \hline
      & \bf ImageNet  &\bf ImageNet-C &\bf ImageNetV2-C\\
    \hline
     Baseline & \bf 69.2 & 32.9 & 24.4\\
     ColorJitter & 68.8 &  33.2 & 24.9 \\
     LowFreq & 68.0 & 34.8 & 26.2 \\
     MixedFreq & 67.8 & \bf 34.9 & 26.2 \\
     Solarize &  68.8 & 33.8 & 25.4 \\
     AutoAug & 68.8 & 34.2 & 25.6 \\
     RandAug & 68.5 & 34.7 & 26.0 \\
     Contrast+Noise & 68.4 & \bf 34.9 & \bf 26.3 \\
     \hline
    \end{tabular}}
    \label{table s9}
\end{table}

\subsection{Choice of Similarity Metrics}
\label{subsec: choice of similarity}
We evaluate the effectiveness of various similarity metrics used in Eq.~(6), including the $l_1$ norm, $l_2$ norm, SVCCA~\cite{raghu2017svcca}, and CKA~\cite{kornblith2019similarity}. As shown in \cref{table s4}, all metrics improve corruption robustness compared to the baseline, demonstrating the flexibility of our method with respect to the choice of similarity measure. While $l_1$ and $l_2$ norms yield slightly higher robustness on ImageNet-C, they also lead to a more significant drop in clean accuracy on ImageNet. In contrast, CKA achieves a better trade-off between robustness and clean performance. Therefore, we adopt CKA as the similarity metric in our main experiments to balance accuracy on both clean and corrupted data.
\begin{table}[h]
    \centering
    \caption{The results of different similarity metrics using ResNet-18 for ImageNet, ImageNet-C, and ImageNetV2-C. Results are reported in top-1 accuracy (\%), with the highest values indicated in \textbf{bold}.}
    \small
    \resizebox{0.95\linewidth}{!}{
    %\fontsize{8}{10}\selectfont
    \begin{tabular}{cccc}
    \hline
      & \bf ImageNet  &\bf ImageNet-C &\bf ImageNetV2-C\\
    \hline
     Baseline & 69.2 & 32.9 & 24.4\\
     $l_1$ & 67.4 & \bf 36.8 & \bf 27.8\\
     $l_2$ & 67.3 & 36.4 & 27.6 \\
     SVCCA & \bf 69.6 & 33.2 & 24.8 \\
     CKA & 68.6 & 34.8 & 26.2 \\
     \hline
    \end{tabular}}
    \label{table s4}
\end{table}

\subsection{Choice of TopK}
\label{subsec: TopK}
We conduct an ablation study on the TopK selection ratio in Eq.~(7), evaluating settings of 5\%, 10\%, 15\%, 30\%, 40\%, and 50\%. As shown in \cref{table s5}, selecting TopK in the range of 30\% to 50\% yields a better trade-off between clean accuracy and corruption robustness, with 50\% (i.e., retaining half the pathways) serving as a balanced and effective default in our framework.

\begin{table}[h]
    \centering
    \caption{The results of different settings on Top-$K$ using ResNet-18 for ImageNet, ImageNet-C and ImageNetV2-C. Results are reported in top-1 accuracy (\%), with the highest values indicated in \textbf{bold}.}
    \small
    \resizebox{0.95\linewidth}{!}{
    %\fontsize{8}{10}\selectfont
    \begin{tabular}{ccccc}
    \hline
      & \bf ImageNet  &\bf ImageNet-C &\bf ImageNetV2-C   \\
    \hline
     Baseline & \textbf{69.2} & 32.9 & 24.4\\
     5\% & 59.0 & 28.2 & 21.3\\
     10\% & 62.4  & 31.1 & 23.2  \\
     15\% & 66.0 & 34.6 & 26.1  \\
     30\% & 68.4 & 35.3 & 26.6  \\
     40\% & 68.6 & \textbf{35.4} & \bf 26.7   \\
     50\% & 69.1 & 34.3 & 25.9  \\
     \hline
    \end{tabular}}
    \label{table s5}
\end{table}

\subsection{Position for Applying S\&D}
\label{subsec: position}
As highlighted in Sec.~3 of our main work, the S\&D refinement can be flexibly applied to any sub-network of a deep model. To investigate how its placement affects performance, we integrate S\&D at different depths of ResNet-18, specifically after the sub-networks defined by $f_{l\leq1}$, $f_{l\leq3}$, $f_{l\leq5}$, $f_{l\leq7}$, and $f_{l\leq9}$. We evaluate these variants on ImageNet, ImageNet-C, and ImageNetV2-C, with results reported in \cref{table s10}. The findings show that applying S\&D at the shallowest position ($f_{l\leq1}$) yields the highest corruption robustness, while positions $f_{l\leq3}$ and $f_{l\leq5}$ still provide meaningful improvements. However, as S\&D is placed deeper in the network, such as at $f_{l\leq7}$ or $f_{l\leq9}$, the robustness gain gradually diminishes. This trend demonstrates that refining pathways early, before non-robust features propagate through subsequent layers, is essential for achieving optimal robustness.

\begin{table}[h]
    \centering
    \caption{Accuracy of ResNet-18 when applying S\&D at varying sub-network depths on ImageNet, ImageNet-C, and ImageNetV2-C. Results are reported in top-1 accuracy (\%), with the highest values indicated in \textbf{bold}.}
    \small
    \resizebox{0.95\linewidth}{!}{
    %\fontsize{8}{10}\selectfont
    \begin{tabular}{cccc}
    \hline
      & \bf ImageNet  &\bf ImageNet-C &\bf ImageNetV2-C\\
    \hline
     Baseline & \bf 69.2 & 32.9 & 24.4\\
     $f_{l\leq1}$ & 68.6 & \bf 34.8 & \bf 26.2\\
     $f_{l\leq3}$ & 69.1 & 33.2 & 24.7 \\
     $f_{l\leq5}$ &  68.6 & 33.0 & 24.8 \\
     $f_{l\leq7}$ & 67.7 & 32.1 & 24.0 \\
     $f_{l\leq9}$ & 66.5 & 30.6 & 22.7 \\
     \hline
    \end{tabular}}
    \label{table s10}
\end{table}

\subsection{Sensitivity to Random Seeds}
To assess the stability of S\&D, we repeat the experiments across multiple random seeds. The results, summarized in \cref{table s6}, show consistent improvements in corruption robustness regardless of the seed, with minimal variance in both clean and corrupted accuracy. This indicates that the performance gains introduced by S\&D are stable and not attributable to favorable random initialization.
\begin{table}[h]
    \centering
    \caption{The results of different random seeds using ResNet-18+S\&D for ImageNet, ImageNet-C, and ImageNetV2-C. \# denotes the seed index. Results are reported in top-1 accuracy (\%), with the highest values indicated in \textbf{bold}.}
    \small
    \resizebox{0.95\linewidth}{!}{
    %\fontsize{8}{10}\selectfont
    \begin{tabular}{cccc}
    \hline
      & \bf ImageNet  &\bf ImageNet-C &\bf ImageNetV2-C\\
    \hline
     Baseline & \bf 69.2 & 32.9 & 24.4\\
     \#1 & 68.5 & 34.2 & 25.8\\
     \#2 & 68.4 &  35.1 &  26.5\\
     \#3 & 68.2 & 34.9 & 26.3 \\
     \#4 & 68.5 & 35.3 & 26.6 \\
     \#5 &  68.3 & 35.4 & 26.6 \\
     \#6 & 68.2 & 35.0 & 26.4 \\
     \#7 &  68.2 & 35.8 & 27.0 \\
     \#8 & 68.5 & \bf 37.6 & \bf 28.9 \\
     \hline
    \end{tabular}}
    \label{table s6}
\end{table}

\subsection{Sensitivity to Learning Rate}
As discussed in Sec.~3.3.3 of our main work, the optimal pathway group $G^*$ produced by S\&D is integrated into the neural network for standard training. Here, we evaluate different learning rates applied to $G^*$ during this phase. Results in \cref{table s7} show that, compared to the baseline, all tested learning rates improve corruption robustness, albeit with a slight drop in clean accuracy. In our main experiments, we fix the learning rate for $G^*$ at 0.0 to preserve the refined robust features without further adaptation.

\begin{table}[h]
    \centering
    \caption{The results of different learning rates using ResNet-18+S\&D for ImageNet, ImageNet-C, and ImageNetV2-C. Results are reported in top-1 accuracy (\%), with the highest values indicated in \textbf{bold}.}
    \small
    \resizebox{0.95\linewidth}{!}{
    %\fontsize{8}{10}\selectfont
    \begin{tabular}{cccc}
    \hline
      & \bf ImageNet  &\bf ImageNet-C &\bf ImageNetV2-C\\
    \hline
     Baseline & \bf 69.2 & 32.9 & 24.4\\
     0.0 & 68.6 & 34.8 & \bf 26.2\\
     0.001 & 67.6 & \bf 35.0 & \bf 26.2\\
     0.01 & 67.9 & 34.3 & 25.6 \\
     0.1 & 68.4 & 34.4 & 25.6 \\
     1 &  68.8 & 34.5 & 25.6 \\
     \hline
    \end{tabular}}
    \label{table s7}
\end{table}

\section{Extended Experimental Results}
\label{sec:results}

\subsection{Extended Results on Image Classification}
\label{subsec: results on image classification}
We present additional image classification results in \cref{table s11} and~\cref{table s12}. \cref{table s11} reports performance on corruption benchmarks using various convolutional neural networks not included in the main text, showing that S\&D consistently improves robustness across various settings. 

\subsection{Fine-Grained Corruption Robustness}
\label{subsubsec:fine_grained_corruption}
We provide a per-corruption breakdown of ImageNet-C performance for MobileNetV2, ResNet-50, and ConvNeXt-base in \cref{table s14}. This complements the averaged robustness results and highlights how S\&D performs across individual corruption types. S\&D consistently improves robustness across most corruptions: all 15/15 corruption types on MobileNetV2, 12/15 on ResNet-50, and 13/15 on ConvNeXt-base, totaling 40 out of 45 architecture-corruption pairs. Gains span noise, blur, weather, and digital corruptions, confirming the method is not overfitting to specific corruptions. Some exceptions exist, such as Snow, Fog, and Brightness on ResNet-50, and Elastic and Pixelate on ConvNeXt-base, indicating areas for further investigation. Overall, this analysis demonstrates that S\&D provides broad, architecture-agnostic improvements in corruption robustness, while also identifying specific corruptions where improvements are limited.

\section{Additional Visualizations}
\label{sec:visualization}

%----------------------------------
\subsection{Extended Visualizations of Robust and Non-robust Features}
\label{subsec: extended visualization}
\begin{figure*}[!ht]
    \centering
    \includegraphics[width=1.0\textwidth]{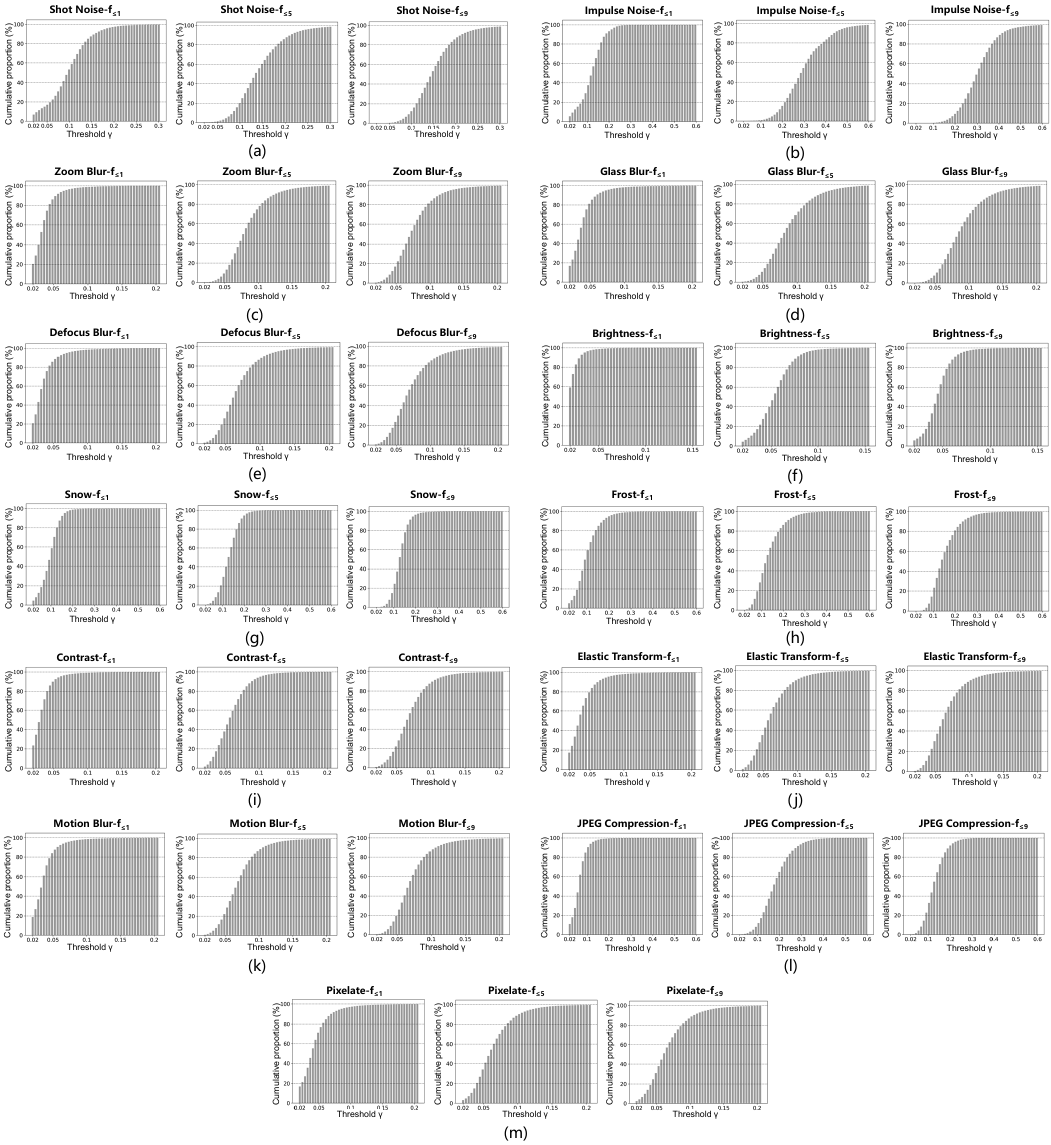}
    \caption{The cumulative distribution of $\gamma$-robust features at different sub-networks under the corruption of (a)~Shot Noise, (b)~Impulse Noise, (c)~Zoom Blur, (d)~Glass Blur, (e)~Defocus Blur, (f)~Brightness, (g)~Snow, (h)~Frost, (i)~Contrast, (j)~Elastic Transform and (k)~Pixelate.}
    \label{Figure 2_extend}
\end{figure*}

%\textbf{Extended visualizations of robust and non-robust features.}
In Sec.~3.2.2, we present the cumulative distributions of $\gamma$-robust features across sub-networks under two common corruptions, including Gaussian Noise and Fog. Here, we extend this analysis by providing visualization results for an additional 13 corruption types, as shown in \cref{Figure 2_extend}. Notably, the patterns observed under these 13 corruptions align closely with those reported in the main work, further corroborating our findings.

%------------------------------------------------------------------------

%----------------------------------
\subsection{Fine-grained Analysis of Robust vs. Non-robust Feature Pathways}
\label{subsec: fine-grained analysis}
\begin{figure*}[!ht]
    \centering
    \includegraphics[width=1.0\textwidth]{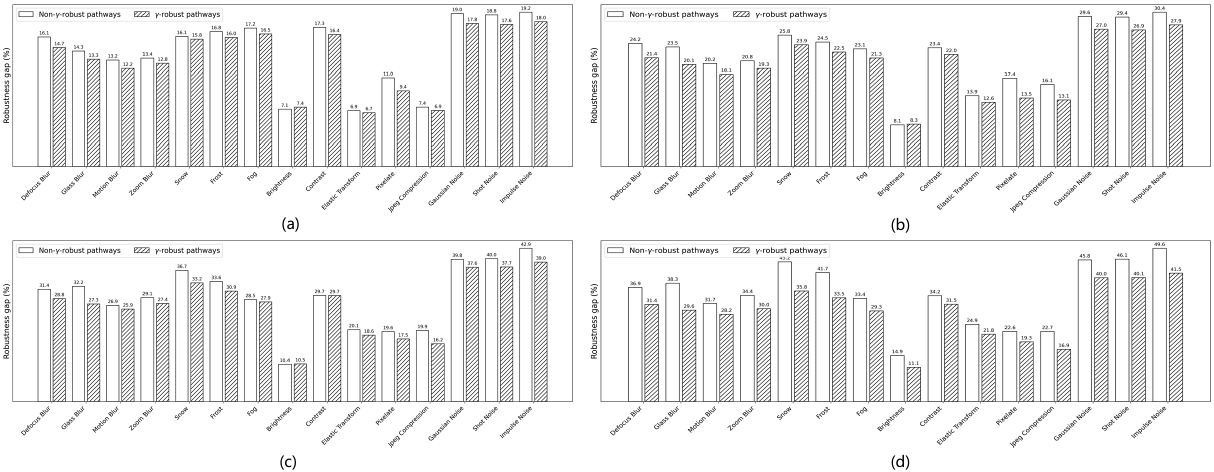}
    \caption{Effective robustness of sub-networks (a) $f_{\leq 1}$, (b) $f_{\leq 5}$, (c) $f_{\leq 9}$, and (d) the full model $f_{\theta}$ when preserving either $\gamma$-robust or non-$\gamma$-robust pathways across 15 corruption types. Lower effective robustness indicates better performance.}
    \label{Figure 3_extend}
\end{figure*}

%\textbf{.}
In Sec~3.2.3, we analyze the impact of selectively retaining $\gamma$-robust versus non-$\gamma$-robust pathways on model robustness under image corruptions, as summarized in Fig.~3 of the main work. Here, we provide extended results for this analysis across all 15 corruption types, as shown in \cref{Figure 3_extend}. The consistent advantage of $\gamma$-robust pathways in enhancing effective robustness is evident across all corruption types, further reinforcing the critical role of robust feature aggregation in the robustness of downstream sub-networks and the full model.

%------------------------------------------------------------------------
\subsection{The Visualization on Semantic Segmentation}
\label{subsec: semantic segmentation}
\begin{figure*}[!ht]
    \centering
    \includegraphics[width=0.9\linewidth]{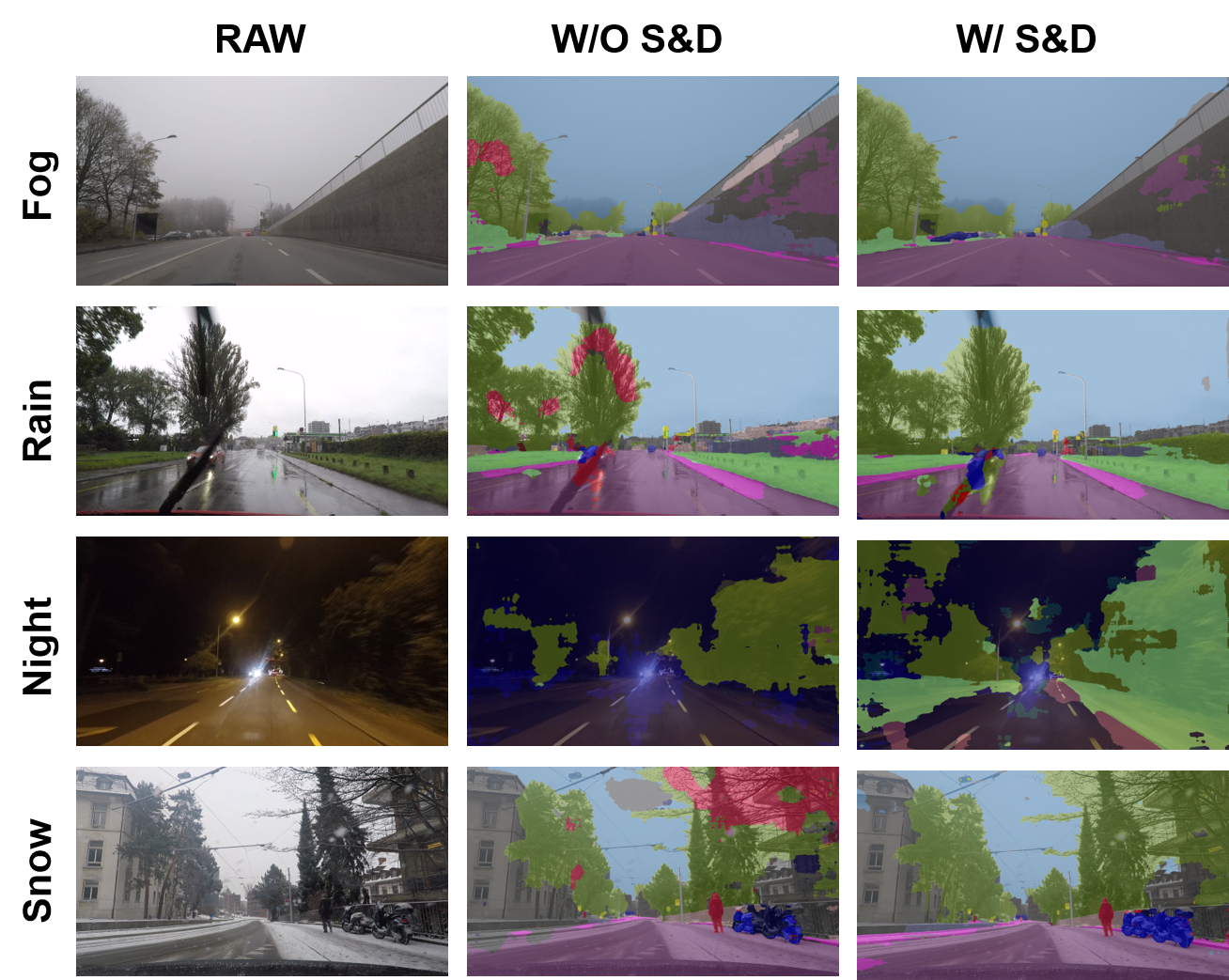}
    \caption{The visualization of segmentation results for the ACDC dataset. }
    \label{Figure 8}
\end{figure*}

%\textbf{The visualization on semantic segmentation.} 
We further test S\&D on the ACDC~\cite{sakaridis2021acdc} dataset, a semantic segmentation under real-world adverse weather conditions, including Fog, Rain, Night, and Snow. The visualization of segmentation results using GCNet~\cite{cao2019gcnet}~(with and without S\&D) is shown in \cref{Figure 8}. S\&D reduces the segmentation errors across different adverse weather scenarios.

\begin{table*}[t]
    \centering
    \caption{We report Top-1 Accuracy (ACC) for ImageNet and mean Corruption Error (mCE) for its corrupted variants (C, $\bar{\text{C}}$, 3DCC, and V2-C). Avg.mCE averages these four error rates. $[\cdot]$ indicates ensemble-based results; $(+)$ and $(-)$ denote performance changes via S\&D. $^*$ indicates advanced training recipes.}
    \small
    \resizebox{0.95\textwidth}{!}{
    \begin{tabular}{ccccccccc}
    \hline
    \textbf{Main} & \bf S\&D  & \bf ImageNet &\bf ImageNet-C & \bf ImageNet-$\bar{C}$ &\bf ImageNet-3DCC &\bf ImageNetV2-C & Avg.mCE~($\downarrow$)\\
    \hline
    \multicolumn{8}{c}{Backbones of Neural Network}\\
    %\multirow{2}{*}{SqueezeNet1\_1} & \ding{55}  & 57.3 & 18.0 & 24.7 & 28.0 & 13.0 & 20.9 \\
    %  &\cellcolor{gray!20} \ding{52}  & \cellcolor{gray!20} 58.3~[57.9] &\cellcolor{gray!20} 19.6 &\cellcolor{gray!20} 25.3 &\cellcolor{gray!20} 28.9 &\cellcolor{gray!20} 14.2 &\cellcolor{gray!20}22.0~(\textbf{+1.1}) \\
     \multirow{2}{*}{AlexNet} & \ding{55}  & 55.6 & 100.0 & 100.0 & 100.0 & 100.0 & 100.0 \\
      &\cellcolor{gray!20} \ding{52}  & \cellcolor{gray!20} 55.1~[55.4] &\cellcolor{gray!20}98.2  &\cellcolor{gray!20}99.8 &\cellcolor{gray!20}99.4  &\cellcolor{gray!20}98.5  &\cellcolor{gray!20}99.0~(\textbf{-1.0})  \\
     \multirow{2}{*}{SqueezeNet1\_0} & \ding{55}  & 57.1 & 103.5 & 101.6 & 100.4 & 102.2 & 101.9 \\
      &\cellcolor{gray!20} \ding{52}  & \cellcolor{gray!20} 57.9~[57.3] &\cellcolor{gray!20}100.1  &\cellcolor{gray!20}99.7  &\cellcolor{gray!20}98.1  &\cellcolor{gray!20} 99.6  &\cellcolor{gray!20}99.4~(\textbf{-1.5})  \\
     \multirow{2}{*}{SqueezeNet1\_1} & \ding{55}  & 57.3 & 104.4 & 101.5 & 100.3 & 102.7 & 102.2 \\
      &\cellcolor{gray!20} \ding{52}  & \cellcolor{gray!20} 58.3~[57.9] &\cellcolor{gray!20} 102.2  &\cellcolor{gray!20}  100.7&\cellcolor{gray!20} 99.0 &\cellcolor{gray!20} 101.1 &\cellcolor{gray!20}100.8~(\textbf{-1.4})  \\
     \multirow{2}{*}{ResNet-18} & \ding{55}  & 69.2 & 84.7 & 87.0 & 82.1 & 88.9 & 85.7 \\
      &\cellcolor{gray!20} \ding{52}  & \cellcolor{gray!20} 68.6~[69.2] &\cellcolor{gray!20} 82.4  &\cellcolor{gray!20} 86.1 &\cellcolor{gray!20} 81.4 &\cellcolor{gray!20} 86.8 &\cellcolor{gray!20}84.2~(\textbf{-1.5})  \\
     \multirow{2}{*}{ResNet-34} & \ding{55}  & 72.8 & 77.9 & 81.4 & 76.3 & 83.8 & 79.9 \\
      &\cellcolor{gray!20} \ding{52}  & \cellcolor{gray!20} 71.9~[72.8] &\cellcolor{gray!20} 74.6 &\cellcolor{gray!20} 80.1 &\cellcolor{gray!20} 74.7 &\cellcolor{gray!20} 81.0 &\cellcolor{gray!20}77.6~(\textbf{-2.3})  \\
     \multirow{2}{*}{ResNet-101} & \ding{55}  & 77.0 & 70.4 & 74.1 & 68.6 & 77.8 & 72.7 \\
      &\cellcolor{gray!20} \ding{52}  & \cellcolor{gray!20} 76.0~[77.0] &\cellcolor{gray!20} 66.7 &\cellcolor{gray!20} 74.3  &\cellcolor{gray!20} 67.0 &\cellcolor{gray!20} 74.7 &\cellcolor{gray!20}70.7~(\textbf{-2.0})  \\
     \multirow{2}{*}{VGG16} & \ding{55}  & 73.3 & 84.6 & 85.8 & 80.7 & 88.7 & 85.0\\
      &\cellcolor{gray!20} \ding{52}  & \cellcolor{gray!20} 75.1~[73.9] &\cellcolor{gray!20} 82.9 &\cellcolor{gray!20} 84.0 &\cellcolor{gray!20} 79.2 &\cellcolor{gray!20} 87.5 &\cellcolor{gray!20}83.4~(\textbf{-1.6})  \\
     \multirow{2}{*}{ResNeXt-50} & \ding{55}  & 77.4 & 72.3 & 74.6 & 69.8 & 79.6 & 74.1\\
      &\cellcolor{gray!20} \ding{52}  & \cellcolor{gray!20} 76.1~[77.4] &\cellcolor{gray!20} 70.8 &\cellcolor{gray!20} 75.3  &\cellcolor{gray!20} 70.0 &\cellcolor{gray!20} 77.9 &\cellcolor{gray!20}73.5~(\textbf{-0.6}) \\
     \multirow{2}{*}{ResNeXt-101} & \ding{55}  & 79.1 & 66.7 & 69.4 & 65.1 & 74.9 &  69.0 \\
      &\cellcolor{gray!20} \ding{52}  & \cellcolor{gray!20} 77.4~[79.1] &\cellcolor{gray!20} 63.0 &\cellcolor{gray!20} 70.2 &\cellcolor{gray!20} 64.2  &\cellcolor{gray!20} 71.7 &\cellcolor{gray!20}67.3~(\textbf{-1.7})  \\
     \multirow{2}{*}{WideResNet-101\_2} & \ding{55}  & 78.8 & 67.7 & 71.8 & 66.1 & 76.2 & 70.5\\
      &\cellcolor{gray!20} \ding{52}  & \cellcolor{gray!20} 77.1~[77.8] &\cellcolor{gray!20} 65.7 &\cellcolor{gray!20} 72.0 &\cellcolor{gray!20} 65.7 &\cellcolor{gray!20} 74.0 &\cellcolor{gray!20}69.4~(\textbf{-1.1})  \\
    \hline
    \multicolumn{8}{c}{Data Augmentation \& Model Regularization}\\
     \multirow{2}{*}{Random Erasing} & \ding{55}  & 76.4 & 76.6 & 78.3 & 72.4 & 82.4 & 77.4 \\
      &\cellcolor{gray!20} \ding{52}  & \cellcolor{gray!20} 74.8~[76.4] &\cellcolor{gray!20} 72.9 &\cellcolor{gray!20} 78.3 &\cellcolor{gray!20} 71.6 &\cellcolor{gray!20} 79.6 &\cellcolor{gray!20}75.6~(\textbf{-1.8}) \\
     \multirow{2}{*}{AutoAug} & \ding{55}  & 76.4 & 73.2 & 76.9 & 69.6 & 79.8 & 74.9\\
      &\cellcolor{gray!20} \ding{52}  & \cellcolor{gray!20} 74.9~[76.4] &\cellcolor{gray!20} 68.6 &\cellcolor{gray!20} 76.7 &\cellcolor{gray!20} 67.9 &\cellcolor{gray!20} 76.1 &\cellcolor{gray!20}72.3~(\textbf{-2.6})  \\
     \multirow{2}{*}{RandAug} & \ding{55}  & 76.5 & 73.7 & 75.6 & 70.3 & 80.1 & 74.9 \\
      &\cellcolor{gray!20} \ding{52}  & \cellcolor{gray!20} 75.1~[76.3] &\cellcolor{gray!20} 68.8 &\cellcolor{gray!20} 76.5 &\cellcolor{gray!20} 68.1 &\cellcolor{gray!20} 76.4 &\cellcolor{gray!20}72.5~(\textbf{-2.4}) \\
     \multirow{2}{*}{CutMix} & \ding{55}  & 76.9 & 76.9 & 76.4 & 72.5 & 82.5 & 77.1 \\
      &\cellcolor{gray!20} \ding{52}  & \cellcolor{gray!20} 75.1~[76.9] &\cellcolor{gray!20} 72.8 &\cellcolor{gray!20} 76.4 &\cellcolor{gray!20} 71.0 &\cellcolor{gray!20} 79.2 &\cellcolor{gray!20}74.9~(\textbf{-2.2}) \\
    \hline
    \multicolumn{8}{c}{Advanced Training Recipe}\\
     \multirow{2}{*}{ResNeXt-50$^*$} & \ding{55}  & 81.1 & 62.6 & 62.4 & 59.8 & 71.3 & 64.0 \\
    &\cellcolor{gray!20} \ding{52}  & \cellcolor{gray!20} 78.7~[79.0] &\cellcolor{gray!20} 60.1 &\cellcolor{gray!20} 63.5 &\cellcolor{gray!20} 61.5 &\cellcolor{gray!20} 69.8 &\cellcolor{gray!20}63.7~(\textbf{-0.3}) \\
     \multirow{2}{*}{ConvNeXt-tiny$^*$} & \ding{55}  & 82.5 & 60.9 & 58.3 & 59.8 & 69.5 & 62.1 \\
    &\cellcolor{gray!20} \ding{52}  & \cellcolor{gray!20} 81.8~[82.1] &\cellcolor{gray!20} 57.0 &\cellcolor{gray!20} 57.0 &\cellcolor{gray!20} 57.8 &\cellcolor{gray!20} 66.7 &\cellcolor{gray!20}59.6~(\textbf{-2.5}) \\
     \multirow{2}{*}{ConvNeXt-small$^*$} & \ding{55}  & 83.5 & 56.2 & 55.7 & 56.0 & 65.8 & 58.4 \\
      &\cellcolor{gray!20} \ding{52}  & \cellcolor{gray!20} 82.6~[82.9] 
    &\cellcolor{gray!20} 53.5 &\cellcolor{gray!20} 54.0 &\cellcolor{gray!20} 55.0 &\cellcolor{gray!20} 63.8 &\cellcolor{gray!20} 56.6~(\textbf{-1.8})\\
    \hline
    \end{tabular}}
    \label{table s11}
\end{table*}

\begin{table*}[t]
    \centering
    \caption{We report Top-1 Accuracy (ACC) for ImageNet and mean Corruption Error (mCE) for its corrupted variants (ImageNet-C with 15 corruption types). Avg.mCE averages these 15 error rates. $[\cdot]$ indicates ensemble-based results; $(+)$ and $(-)$ denote performance changes via S\&D. $^*$ indicates advanced training recipes.}
    \small
    \resizebox{1.0\textwidth}{!}{
    \begin{tabular}{cccccccccccccccccc}
    \hline
    \textbf{Main} & \bf S\&D & \bf ImageNet & \bf Gauss. & \bf Shot & \bf Impulse & \bf Defocus & \bf Glass & \bf Motion & \bf Zoom & \bf Snow & \bf Frost & \bf Fog & \bf Bright & \bf Contrast & \bf Elastic & \bf Pixel & \bf JPEG \\
    \hline
     \multirow{2}{*}{MobileNetV2$^*$} & \ding{55}  & 72.1 & 89.7 & 89.9 & 86.5 & 87.4 & 96.9 & 86.7 & 88.5 & 82.3 & 83.3 & 67.8 & 65.7 & 64.7 & 93.8 & 115.6 & 92.6 \\
    & \cellcolor{gray!20} \ding{52}  & \cellcolor{gray!20} 72.5~[72.3] & \cellcolor{gray!20} 76.8 & \cellcolor{gray!20} 77.2 & \cellcolor{gray!20} 78.2 & \cellcolor{gray!20} 82.5 & \cellcolor{gray!20} 92.7 & \cellcolor{gray!20} 81.7 & \cellcolor{gray!20} 87.1 & \cellcolor{gray!20} 80.6 & \cellcolor{gray!20} 75.3 & \cellcolor{gray!20} 65.9 & \cellcolor{gray!20} 65.0 & \cellcolor{gray!20} 61.4 & \cellcolor{gray!20} 90.8 & \cellcolor{gray!20} 81.1 & \cellcolor{gray!20} 81.2\\
    \hline
     \multirow{2}{*}{ResNet-50$^*$} & \ding{55}  & 80.7 & 67.5 & 68.3 & 68.7 & 67.5 & 85.9 & 73.8 & 72.0 & 64.6 & 62.0 & 48.8 & 46.4 & 46.3 & 79.5 & 80.7 & 65.4  \\
    & \cellcolor{gray!20} \ding{52}  & \cellcolor{gray!20} 79.5~[79.7]  & \cellcolor{gray!20} 57.0 & \cellcolor{gray!20} 57.1 & \cellcolor{gray!20} 57.4 & \cellcolor{gray!20} 65.1 & \cellcolor{gray!20} 78.5 & \cellcolor{gray!20} 69.0 & \cellcolor{gray!20} 71.2 & \cellcolor{gray!20} 67.0 & \cellcolor{gray!20} 58.2 & \cellcolor{gray!20} 52.5 & \cellcolor{gray!20} 50.2 & \cellcolor{gray!20} 45.9 & \cellcolor{gray!20} 75.5 & \cellcolor{gray!20} 54.7 & \cellcolor{gray!20} 63.7\\
    \hline
     \multirow{2}{*}{ConvNeXt-base$^*$} & \ding{55}  & 84.0 & 50.0 & 51.6 & 49.6 & 60.1 & 72.6 & 53.6 & 63.3 & 50.2 & 46.8 & 50.8 & 39.4 & 39.4 & 63.0 & 58.6 & 54.6 \\
    & \cellcolor{gray!20} \ding{52}  & \cellcolor{gray!20} 82.9~[83.4] & \cellcolor{gray!20} 44.8 & \cellcolor{gray!20} 45.0 & \cellcolor{gray!20} 44.2 & \cellcolor{gray!20} 57.4 & \cellcolor{gray!20} 69.0 & \cellcolor{gray!20} 51.9 & \cellcolor{gray!20} 60.6 & \cellcolor{gray!20} 47.0 & \cellcolor{gray!20} 43.1 & \cellcolor{gray!20} 44.3 & \cellcolor{gray!20} 38.4 & \cellcolor{gray!20} 37.0 & \cellcolor{gray!20} 64.6 & \cellcolor{gray!20} 60.1 & \cellcolor{gray!20} 53.8\\
    \hline
    \end{tabular}}
    \label{table s14}
\end{table*}


%You can have as much text here as you want. The main body must be at most $8$
%pages long. For the final version, one more page can be added. If you want, you
%can use an appendix like this one.

%The $\mathtt{\backslash onecolumn}$ command above can be kept in place if you
%prefer a one-column appendix, or can be removed if you prefer a two-column
%appendix.  Apart from this possible change, the style (font size, spacing,
%margins, page numbering, etc.) should be kept the same as the main body.
%%%%%%%%%%%%%%%%%%%%%%%%%%%%%%%%%%%%%%%%%%%%%%%%%%%%%%%%%%%%%%%%%%%%%%%%%%%%%%%
%%%%%%%%%%%%%%%%%%%%%%%%%%%%%%%%%%%%%%%%%%%%%%%%%%%%%%%%%%%%%%%%%%%%%%%%%%%%%%%

\end{document}